\documentclass{article}

\PassOptionsToPackage{numbers, compress}{natbib}

\usepackage[preprint]{neurips_2024}

\usepackage[utf8]{inputenc} 
\usepackage[T1]{fontenc}    
\usepackage{hyperref}       
\usepackage{url}            
\usepackage{booktabs}       
\usepackage{amsfonts}       
\usepackage{nicefrac}       
\usepackage{microtype}      
\usepackage{xcolor}         

\usepackage{graphicx}
\usepackage{multirow}
\usepackage{multicol,enumitem,lineno}
\usepackage{wrapfig}
\usepackage{algpseudocode}
\usepackage{algorithm}

\usepackage{amsmath}
\usepackage{amssymb}
\usepackage{mathtools}
\usepackage{amsthm}
\usepackage{xcolor,colortbl}
\usepackage{pgfplots}
\pgfplotsset{compat=newest}
\pgfplotsset{scaled y ticks=false}
\usepgfplotslibrary{groupplots}
\usepgfplotslibrary{dateplot}
\usepackage{tikz}
\usepackage{xcolor}

\usepackage[capitalize,noabbrev]{cleveref}

\theoremstyle{plain}
\newtheorem{theorem}{Theorem}[section]
\newtheorem*{theorem*}{Theorem}
\newtheorem{proposition}[theorem]{Proposition}

\newtheorem{corollary}[theorem]{Corollary}
\theoremstyle{definition}
\newtheorem{definition}[theorem]{Definition}

\theoremstyle{remark}
\newtheorem{remark}[theorem]{Remark}

\title{CoNO: Complex Neural Operator for Continous Dynamical Physical Systems}

\newcommand{\cono}{\textsc{CoNO}{}}
\author{%
  Karn Tiwari\\
  Department of Electrical Communication Engineering\\
  Indian Institute of Science, Bangalore\\
  Bengaluru, 560012, India \\
  \texttt{karntiwari@iisc.ac.in} \\
  \And
  N M Anoop Krishnan\\
  Yardi School of Artificial Intelligence\\
  Indian Institute of Technology, Delhi\\
  New Delhi, 110016, India \\
  \texttt{krishnan@iitd.ac.in} \\
  \And
  Prathosh A P\\
  Department of Electrical Communication Engineering\\
  Indian Institute of Science, Bangalore\\
  Bengaluru, 560012, India \\
  \texttt{prathosh@iisc.ac.in} \\
}

\begin{document}

\maketitle

\begin{abstract}
Neural operators extend data-driven models to map between infinite-dimensional functional spaces. While these operators perform effectively in either the time or frequency domain, their performance may be limited when applied to non-stationary spatial or temporal signals whose frequency characteristics change with time. 
Here, we introduce a Complex Neural Operator (\cono{}) that parameterizes the integral kernel using Fractional Fourier Transform (FrFT), better representing non-stationary signals in a complex-valued domain. Theoretically, we prove the universal approximation capability of \cono{}. We perform an extensive empirical evaluation of \cono{} on seven challenging partial differential equations (PDEs), including regular grids, structured meshes, and point clouds. Empirically, \cono{} consistently attains state-of-the-art performance, showcasing an average relative gain of 10.9\%. Further, \cono{} exhibits superior performance, outperforming all other models in additional tasks such as zero-shot super-resolution and robustness to noise. \cono{} also exhibits the ability to learn from small amounts of data---giving the same performance as the next best model with just 60\% of the training data. Altogether, \cono{} presents a robust and superior model for modeling continuous dynamical systems, providing a fillip to scientific machine learning. 
\end{abstract}
\vspace{-0.5cm}
\section{Introduction}
\vspace{-0.2cm}
Continuum systems span various scientific and engineering fields, such as robotics, biological systems, climate modeling, and fluid dynamics, among others \cite{debnath2005nonlinear}. These systems are represented using Partial Differential Equations (PDEs), the solution of which provides the system's time evolution. The solution of PDEs necessitates the identification of an optimal solution operator, which maps across functional spaces while including the initial conditions and coefficients. Traditionally, numerical methods, such as finite element and spectral methods, have been employed to approximate the solution operator for PDEs. However, these approaches often incur high computational costs and exhibit limited adaptability to arbitrary resolutions and geometries ~\citep{sewell2012analysis}. Such high computational costs of these numerical methods inhibit the real-time prediction crucial in weather forecasting and robotics.

\begin{figure}[!htb]
    \centering
    \includegraphics[width=1.0\linewidth]{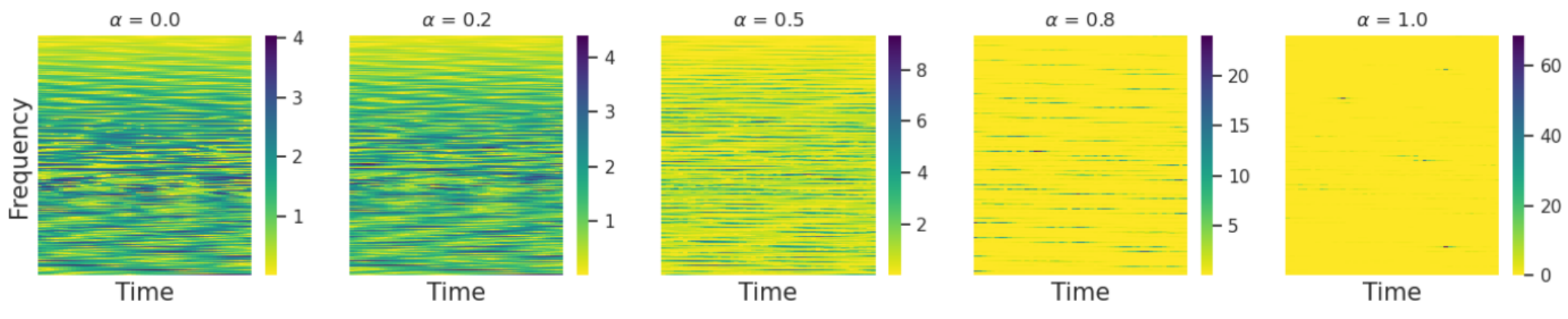}
    \vspace{-0.5cm}
    \caption{\label{fig:figure1} \small FrFT heatmaps illustrating the temporal-frequency characteristics of the 2D Navier-Stokes equation for varying values of $\alpha$. Each subplot represents the magnitude of the transformed frequency content over time, obtained by applying the FrFT and then flattened 2D frequency map of the Navier-Stokes equation. Different subplots correspond to fractional orders $\alpha$, highlighting the diverse spectral behaviors captured by the FrFT across both the temporal and frequency domains. Note that $\alpha = 0$ represents the time domain while $\alpha = 1$ represents the frequency domain.}
    \vspace{-0.5cm}
\end{figure}

Recently, in the realm of scientific machine learning, neural networks present a promising alternative to solve PDEs through a data-driven approach. Specifically, neural operators represent an extension of neural networks, facilitating the mapping between infinite-dimensional functional spaces and serving as a universal approximation of the operator. Notably, these operators learn the functionals without any prior knowledge of the underlying PDE, relying solely on data-driven training, leading to faster inference times than the traditional methods \cite{li2020fourier, kovachki2021neural}. 

Among different neural operators, the Fourier Neural Operator (FNO) \cite{li2020fourier} has gained widespread recognition due to its ability to navigate the infinite-dimensional functional space via kernel integral operations in the Fourier domain. Renowned for its versatility, FNO has found successful applications across diverse domains, including weather forecasting \cite{kurth2023fourcastnet}, biomedical surrogate modeling \cite{guan2021fourier}, and expediting sampling processes in diffusion models \cite{zheng2023fast}. Nevertheless, recent investigations have brought to light some specific challenges associated with FNO, including aliasing errors \cite{fanaskov2022spectral}, a departure from continuous-discrete equivalence \cite{bartolucci2023neural}, susceptibility to vanishing gradient problems with an increasing number of layers \cite{tran2021factorized} and susceptibility to poor performance when presented with noisy data \cite{burark2024codbench}. FNO also exhibits suboptimal performance on time-dependent PDEs \cite{li2023long, burark2024codbench} exemplified by the turbulent Navier-Stokes equations (chaotic flow). Notably, FNO encounters difficulties in making predictions over extended time horizons, constrained by the limitations of the Fourier Transform (FT), which is tailored for stationary signals and needs more time-frequency characteristics---as it decomposes the function based on monochromatic basis function and is unable to detect oscillatory patterns present in chirps signal \cite{shabbir2019approximation}. For instance, the change of frequency with time in the Navier-Stokes equation is non-stationary and nonlinear (see Fig. ~\ref{fig:figure1} and App. Fig. \ref{fig:figure6}), leading to the concentration of spectrum around low frequency in the Fourier Transform. Note that the Navier-Stokes equation holds significant relevance across diverse practical domains, including but not limited to aerodynamics, weather forecasting, and oceanography. Addressing these challenges through a data-driven methodology underscores the pressing need for alternative operator formulation that has ability to learn non-stationary signals.

A generalization of the FT for non-stationary signals, where spectra evolve with time, by employing fractional FT (FrFT) can potentially improve the prediction of long-horizon dynamics, particularly in handling highly nonlinear and rapidly changing time-frequency characteristics \cite{chassande1999time}. FrFT improves the efficiency of the classical FT by using a general parameterized adjustable orthonormal basis. This basis allows us to break down a chirp function into simpler parts. The adjustable parameter helps efficiently reconstruct the chirp functions \cite{shabbir2019approximation}. The FrFT generalizes FT by rotating the signal between the time and frequency domains, transitioning from real to complex domain, and incorporating phase information \cite{ozaktas2001fractional}, resulting in a complex representation of the signal. However, the complex-valued representations have remained unexplored in the operator learning paradigm, although a large literature exists on complex-valued neural networks (CVNNs) \cite{nitta2002critical}. CVNNs offer easier optimization \cite{nitta2002critical}, faster convergence during training \cite{arjovsky2016unitary, danihelka2016associative, wisdom2016full}, better generalization \cite{hirose2012generalization}, data efficiency, and resilience to noise \cite{danihelka2016associative, wisdom2016full}. Thus, combining the operator paradigm with CVNNs can potentially result in an architecture that exploits both best, leading to a model that provides improved long-horizon prediction.

\textbf{Our Contributions.} Motivated by the above observations, we introduce the Complex Neural Operator (\cono{}). The major contributions of the present work are as follows.  
\begin{enumerate}
    \item \textbf{Novel architecture.} \cono{} represents the first instance that performs operator learning employing a CVNN that parameterizes the integral kernel through FrFT, enabling richer information through phase details learnable by CVNNs suitable for non-stationary signals. Table \ref{tab:table1} shows the comprehensive comparison of \cono{} with current SOTA operators.
    \item \textbf{Universal approximation.} We prove theoretically that \cono{} follows the universal approximation theorem for operators (see Thm 4.3).
    \item \textbf{Superior performance.} We show that \cono{} consistently exhibits superior performance compared to SOTA operators, always ranking among the top two on all the datasets with an average gain of 10.9\% as presented in Table ~\ref{tab:experiment_benchmarks}. 
    \item \textbf{Data efficiency and robustness to noise.} \cono{} demonstrates superior performance even with limited samples, data instances, and training epochs. \cono{} provides more robustness when noise is injected into the training data compared to the existing methods and performs better than SOTA methods even under 0.1\% data noise.
\end{enumerate}

\vspace{-0.4cm}
\section{Related Work}
\vspace{-0.3cm}
\begin{wraptable}{r}{7cm}
    \centering
    \caption{\small A comprehensive comparative analysis of features of state-of-the-art operators with \cono{}. "*" denotes not applicable.}
    \label{tab:table1}
    \resizebox{\linewidth}{!}{ 
    \begin{tabular}{l|c|c|c}
        \toprule
        \textbf{Features} & \textbf{FNO} &\textbf{LSM}            & \textbf{\cono{} (Ours)}\\
        \midrule
        \midrule
        Integral Kernel & Frequency & Spatial & Spatial-Frequency\\
        Elementary function & Sine & * & Linear Frequency Modulation \\
        Representation  & Real & Real  & Complex\\
        Pertinent Signal & Stationary & * & Time Varying Signal \\
        $\alpha$ Parameter & Fixed ($90^{o}$) & Fixed ($0^{o}$) & Not Fixed (Learnable) \\
        Applicability & Individual & Individual & Unified \\
        \bottomrule
    \end{tabular}}
\end{wraptable} 

\textbf{Neural Operators (NO):} Neural operators have shown promise in solving PDEs in a data-driven fashion \citep{kovachki2021neural}. \citet{lu2021learning} introduced DeepOnet, theoretically establishing its universal approximation capabilities. The DeepOnet architecture comprises a branch network and a trunk network, with the former dedicated to learning the input function operator and the latter tasked with learning the function space onto which it is projected. Another famous architecture, the FNO, proposed by \citet{li2020fourier}, utilizes a frequency domain method. FNO incorporates Fourier kernel-based integral transformations through fast Fourier transform and projection blocks. An enhanced version, F-FNO \cite{tran2021factorized}, improves upon the original FNO by integrating distinct spectral mixing and residual connections. Subsequently, various kernel integral neural operators based on transformations have emerged. For example, \citet{fanaskov2022spectral} introduced spectral methods, such as Chebyshev and Fourier series, to mitigate aliasing errors and enhance clarity in FNO outputs. \citet{li2022fourier} incorporated specialized learnable transforms to facilitate operator learning on irregular domains, achieved by transforming them into uniform latent meshes. 

\citet{kovachki2021neural} demonstrated that the well-known attention mechanism can be seen as a particular case of neural operator learning the integral kernel specifically applied to address irregular meshes for solving PDEs, a characteristic managed by Geo-FNO \cite{li2022fourier} differently. \citet{cao2021choose} employed two self-attention mechanism-based operators without utilizing a softmax layer, accompanied by a theoretical interpretation. Recently, \citet{hao2023gnot} introduced the GNOT operator, featuring a linear cross-attention block designed to enhance the encoding of irregular geometries. However, transformer-based operators are susceptible to issues arising from limited data samples, displaying a tendency to overfit easily on training data without exhibiting robust generalization. In addressing the challenges posed by multiscale PDEs, \citet{liu2022ht} proposed a hierarchical attention-based operator.

\textbf{Fractional Fourier Transform (FrFT):} 
The FrFT represents a generalization of the classical Fourier Transform (FT), providing robust capabilities for spectral analysis. It achieves this by facilitating the transformation of input signals into an intermediate domain between the time and frequency domains, thereby establishing a time-frequency representation \cite{ozaktas2001fractional}. FrFT is particularly effective in processing non-stationary signals, commonly called "chirp signals" or signals exhibiting frequency derivatives over time. App. Fig. \ref{fig:figure7} illustrates the efficacy of employing the FrFT for noise filtering within the signal spectrum through the rotation of the fractional order axis. In contrast to FrFT, alternative signal processing techniques such as wavelet or Gabor transformation don't provide a joint signal energy distribution across both time and frequency domains. Additionally, these alternatives often grapple with challenges related to high computational complexity, the selection of wavelet functions, and sensitivity to signal noise. FrFT, with its distinct advantages, has found applications in various domains, ranging from solving differential equations \cite{mcbride1987namias}, wireless communication \cite{li2018multi}, biomedical signal processing \cite{gomez2020fractional}, image encryption and image compression \cite{naveen2019lossless} etc. \citet{yu2024deep} demonstrated the benefits of employing FrFT over conventional convolution across various tasks in computer vision, including segmentation, object detection, classification, super-resolution, etc.

\textbf{Complex Valued Neural Networks (CVNNs):} 
A CVNN incorporates complex-valued parameters and variables within its architecture, enabling the representation of both magnitude and phase information in the neural networks \cite{lee2022complex, chiheb2017deep}. The utilization of CVNNs encompasses diverse advantages, extending from biological to signal processing applications. \citet{danihelka2016associative} has demonstrated that complex numbers enhance efficiency and stability in information retrieval processes. Additionally, \citet{arjovsky2016unitary} have introduced complex recurrent neural networks (RNNs), highlighting that unitary matrices offer a more intricate representation, thereby mitigating issues associated with vanishing and exploding gradient problems. In image processing, phase information is a critical descriptor, offering detailed insights about image shape, orientation, and edges. \citet{oppenheim1981importance} showed the information encapsulated in the phase of an image proves sufficient for the recovery of a substantial proportion of the encoded magnitude information. 

CVNNs have been a research focus for a long time \cite{georgiou1992complex, kim2003approximation, hirose2003complex, nitta2004orthogonality}. 
Recently, \citet{geuchen2023universal} established the universal approximation capabilities of CVNNs for deep, narrow architectures, significantly contributing to understanding their expressive power. Prior works \cite{reichert2013neuronal, tygert2016mathematical, arjovsky2016unitary, danihelka2016associative, chatterjee2022complex, geuchen2024optimal} have made noteworthy strides in both experimental and theoretical aspects of CVNNs. In the domain of computer vision, the utilization of scattering transformation-based complex neural networks has demonstrated considerable promise, showcasing their ability to achieve performance on par with real-valued counterparts while employing significantly fewer parameters \cite{ko2022coshnet, worrall2017harmonic, rawat2021novel}. In NLP, complex embeddings have been incorporated for semantic and phonetic processing of natural languages \cite{trouillon2017complex, demir2021convolutional}. In contrast, \citet{yang2020complex} and \citet{dong2021signal} showcased the advantages of employing CVNNs transformers for the NLP community. Despite these notable applications across various domains, exploring the applicability of CVNNs within the SciML community is still limited.

\vspace{-0.3cm}
\section{Preliminaries}
\vspace{-0.3cm}
\textbf{Problem Setting:} We have followed and adopted the notations in \citet{li2020fourier}. Let $D$ denote a bounded open set as $D \subset \mathbb{R}^d$, with $A = A(D;\mathbb{R}^{d_a})$ and $U = U(D;\mathbb{R}^{d_u})$ as separable Banach spaces of functions representing elements in $\mathbb{R}^{d_a}$ and $\mathbb{R}^{d_u}$, respectively. Consider $\mathcal{G}^\dagger: A \rightarrow U$ to be a nonlinear surrogate mapping arising from the solution operator for a parametric PDE. It is assumed that there is access to i.i.d. observations ${(a_j,u_j)}_{j=1}^N$, where $a_j \sim \mu$, drawn from the underlying probability measure $\mu$ supported on $A$, and $u_j = \mathcal{G}^\dagger(a_j)$.

The objective of operator learning is to construct an approximation for $\mathcal{G}^\dagger$ via a parametric mapping $\mathcal{G}: A \times \Theta \rightarrow U$, or equivalently, $\mathcal{G}_\theta: A \rightarrow U, \theta \in \Theta$, within a finite-dimensional parameter space $\Theta$. The aim is to select $\theta^\dagger \in \Theta$ such that $\mathcal{G}(\cdot,\theta^\dagger) = \mathcal{G}_\theta^\dagger \approx \mathcal{G}^\dagger$.  
This framework facilitates learning in infinite dimensional spaces as the solution to the optimization problem in Eq.~\ref{eq1} constructed using a loss function $\mathcal{L}: U \times U \rightarrow \mathbb{R}$. 
\begin{equation}
    \label{eq1}
    \min_{\theta \in \Theta} \mathbb{E}_{a \sim \mu} \left[ \mathcal{L}(\mathcal{G}(a,\theta),\mathcal{G}^\dagger(a)) \right],
\end{equation}
The optimization problem is solved in operator learning frameworks using a data-driven empirical approximation of the loss function akin to the regular supervised learning approach using train-test observations. Usually, $\mathcal{G}_\theta$ is parameterized using deep neural networks. 

\textbf{Fractional Fourier Transform (FrFT): } Inspired by the kernel formulation for solving linear PDEs using Green's function, we construct the model $\mathcal{G}_{\theta}$ employing an iterative approach to map an input function $a$ to an output function $u$ within the \cono{} framework as detailed in Sec.~\ref{sec:cono}. In \cono{}, the kernel integral is formulated using the FrFT with a learnable order. The fractional transformation of order $\alpha$ ($\alpha \in \mathbb{R}$) is a parameter of the Fractional Fourier Transform (FrFT) that corresponds to the $\alpha^{th}$ power of the Fourier Transform (FT) (denote by $\mathcal{F}^{\alpha}$).

\begin{definition} [FrFT]
The fractional Fourier transform with angle $\alpha$ of a signal $f(y)$ is defined as:

\begin{equation}
    \mathcal{F}^{\alpha} (f)(m) = \int_{-\infty}^{\infty} f(y) \mathcal{K}_{\alpha}(m, y) \, dy, \label{eq:frft}
\end{equation}
where, 
\begin{equation*}
    \mathcal{K}_\alpha(m, y) =
    \begin{cases}
        c(\alpha) \exp\{j\pi a(\alpha)[(m^2 + y^2) - 2b(\alpha) m y]\} & \text{if } \alpha \neq  n\pi\mathbb{Z}, \\
        \delta(m - y) & \text{if } \alpha =  2\pi\mathbb{Z}, \\
        \delta(m + y) & \text{if } \alpha + \pi =  2\pi 	\mathbb{Z}
    \end{cases}
\end{equation*}
where, $a(\alpha) = \cot \alpha, \: b(\alpha) = \csc \alpha, \: \text{and} \:  c(\alpha) = \sqrt{1 - j \cot \alpha}$ and  $\mathcal{F}^{\alpha} (f)(m)$ denotes the $m^{th}$ fractional fourier coefficient of order $\alpha$ of $f$.
\end{definition}
\begin{remark}
    For Eq. \ref{eq:frft}, it reduces to \textbf{Standard Fourier Transform (FT)} when $\alpha = \frac{\pi}{2}$, then $\cot{\alpha} = 0$ and $\csc{\alpha} = 1$.
\end{remark}

\textbf{Complex Valued Neural Networks (CVNNs):} In the \cono{} framework, the integration of kernels in the operator $\mathcal{G}_{\theta}$ is performed within the complex-valued domain using CVNNs. A CVNN is modeled as real and imaginary parts or magnitude and phases as follows \cite{trouillon2017complex}:
\begin{equation}
    z = x + jy = |z|e^{j\angle z}
\end{equation}
where $j = \sqrt{-1}$ is imaginary unit, $x$ and $y$ are the real and imaginary parts of $z$, and $|z|$ and $\angle z$ are the magnitude and phase of $z$.

\begin{definition}[Complex Valued Activation]
    Let $z \in \mathbb{C}$ be a complex number with real part $\text{Re}(z)$ and imaginary part $\text{Im}(z)$. The Complex GeLU (CGeLU) activation function is defined as follows:
    \begin{equation}
        \text{CGeLU}(z) = \text{GeLU}(\text{Re}(z)) + j \cdot \text{GeLU}(\text{Im}(z)),
        \label{eq:cgeelu}
    \end{equation}
    where $\text{GeLU}(\cdot)$ is the Gaussian Error Linear Unit activation function \cite{hendrycks2016gaussian}.
    The CGeLU activation function satisfies the Cauchy-Riemann equations when the real and imaginary parts of $z$ are strictly positive or negative.
\end{definition}

\textbf{Complex Valued Back-propagation:} Complex-valued back-propagation involves extending traditional back-propagation algorithms to handle complex numbers, utilizing mathematical tools like Wirtinger calculus \cite{amin2011wirtinger} which enable training neural networks with complex-valued weights and activation, allowing for the modeling of intricate relationships within complex data domains \cite{barrachina2023theory, chiheb2017deep}.

\vspace{-0.3cm}
\section{Complex Neural Operator (\cono{})}
\vspace{-0.3cm}
\label{sec:cono}
\subsection{Proposed Method}
\begin{figure*}[t]
    \centering
    \includegraphics[width=0.9\linewidth]{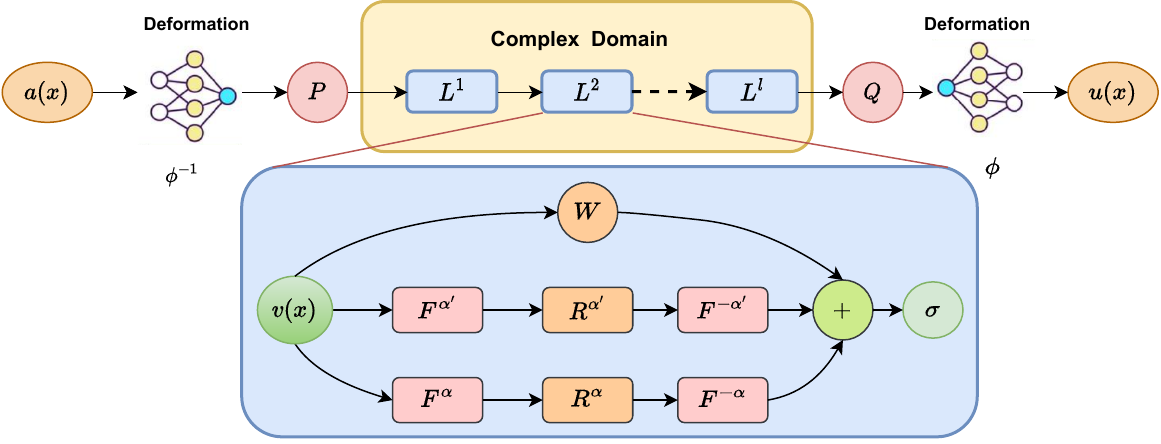}
    \vspace{-0.2cm}
    \caption{\label{fig:cono} \small \textbf{\cono{} Architecture Overview.} \textbf{(Top)} (1) The input function $a(x)$ undergoes a deformation $\phi^{-1}$ to convert an irregular mesh into a uniform mesh. (2) The deformed input is then lifted to a higher dimension in the channel space using a neural network. (3) Apply iterative \cono{} layer in complex domain consisting of fractional integral kernel. (4) Then, the output is projected back on the lower dimension in channel space using neural network. (5) Solution $u(x)$ is obtained by passing through the deformation $\phi$. \textbf{(Bottom)} Zoomed version of FrFT integral kernel defined in Eq. \ref{equ: non linear operator} with learnable parameters $R^{\alpha}$, $R^{\alpha'}$, $W$ and fractional order $\alpha$ and $\alpha'$.} 
\end{figure*}

\vspace{-0.3cm}
Here, we introduce our framework, named the Complex Neural Operator (\cono{}) depicted in Fig. \ref{fig:cono}.

\textbf{Overview:} Suppose $\mathcal{G}_{\theta}$ represents an iterative sequence of functions. The operator $\mathcal{G}_{\theta}:A \rightarrow U$, where $A = A(D; \mathbb{R}^{d_a})$ and $U = U(D; \mathbb{R}^{d_u})$ are separable Banach spaces of functions, representing elements in $\mathbb{R}^{d_a}$ and $\mathbb{R}^{d_u}$, respectively. Our goal is to construct the operator in a structure-preserving manner where we band-limit a function over a given spectrum \cite{vetterlifoundations}, thus preserving complex continuous-discrete equivalence such that Shannon-Whittaker-Kotel’nikov theorem is obeyed for all continuous operations \cite{unser2000sampling}. With this, our operator \cono{}, denoted by $\mathcal{G}_\theta$ is defined as follows:
\begin{equation}
    \mathcal{G}_\theta = \phi \circ \mathcal{Q} \circ \mathcal{L}_l \circ \ldots \circ \ \mathcal{L}_1 \circ \mathcal{P} \circ \phi^{-1},\label{eq:cono}
\end{equation}
where $\circ$ denotes composition. On irregular geometries, $\phi$ represents the function modeled by a neural network that maps the irregular domain into a regular latent mesh for operator learning. The operators $\mathcal{P}: \mathbb{R}^{d_a} \rightarrow \mathbb{R}^{d_1}$ and $\mathcal{Q}: \mathbb{R}^{d_l} \rightarrow \mathbb{R}^{d_u}$ correspond to the lifting and projection operations, encoding lower-dimensional spaces into higher-dimensional spaces or vice versa which helps in converting the non-linear dynamics into linear dynamics inspired by Koopman Operator \cite{bevanda2021koopman}. The operator consists of $l$ layers of non-linear integral operators $\sigma(\mathcal{L}_l)$ (Eq. \ref{equ: non linear operator}), where $\sigma$ is an activation function which is applied from layer $\mathcal{L}_{i}$, where $i$ belongs to $1$ to $l-1$ to introduce non-linearity in operators, akin to standard neural networks, to learn highly nonlinear operators, and $\theta$ denotes all the learnable parameters of the operator.

\textbf{Non-linear Operator Layer, $\mathcal{L}$:} In \cono{}, the complex kernel operator is defined as follows:
\begin{equation}
    v^{l+1}(x) = \sigma\left(Wv^l(x) + b^l + \mathcal{K}(a, \alpha) v^l(x) \right) \quad \forall x \in D \label{equ: non linear operator}
\end{equation}

where: $v^l$ is the representation at layer $l$ and $v^l \in \mathbb{C}^{d_l}$, $ \mathcal{K}:V^{l}\times\Theta_K \to \mathcal{L}(V^{l+1}(D;\mathbb{C}^{d_l}), V^{l+1}(D;\mathbb{C}^{d_{l+1}})) $ maps to bounded linear operators on $V^{l+1}(D;\mathbb{R}^{d_{l+1}})$ as integral operator is linear and bounded operator as shown in App. section \ref{Appendix B},
$\sigma$ is a non-linear complex activation function,
$W$ is a linear transformation, $W: \mathbb{C}^{d_l} \rightarrow \mathbb{C}^{d_{l+1}}$,
$\mathcal{K}(a, \phi)$ is an integral kernel parameterized by CVNN (Eq. \ref{equ: integral kernel}), $a$ is a complex input function, and $\alpha$ belongs to the parameter space of $\mathcal{G}_\theta$ and $b$ is the bias.

\textbf{Fractional Integral Operator, $\mathcal{K}(a, \alpha)$:} In \cono{}, the integral operator is defined as, $\mathcal{K}(a, \alpha): V^l(D) \rightarrow V^l(D)$ as follows:
\begin{equation}
    \mathcal{K}(a, \alpha)v^{l}(x) = \sum_{\alpha, \alpha'} \mathcal{F}^{-\alpha}(R^\alpha \cdot (\mathcal{F}^\alpha v^{l}))(x)  \quad \forall x \in D \label{equ: integral kernel}
\end{equation}
where, $\mathcal{F}^\alpha$ denotes the FrFT with order $\alpha$, $\mathcal{F}^{-\alpha}$ denotes the inverse FrFT, $R^\alpha$ is the learnable function which we learn from data, $a$ is a complex input function, $\alpha$ is the learnable fractional order which we learn with data as well and $\alpha$ belongs to the parameter space of $\mathcal{G}_\theta$. Note that although $R^\alpha$ can be parameterized using a neural network, in the present work, we observed that linear transformation provided slightly better performance than deep neural networks and hence used the same. In the subsequent subsection, we theoretically prove that the proposed operator can map between infinite-dimensional spaces, leading to the learning of non-linear operators.
\vspace{-0.3cm}
\subsection{Theoretical Analysis}
\vspace{-0.2cm}
In this subsection, we present a theoretical analysis of several components of the proposed method. Specifically, we prove the following. (i) The $N$-dimensional FrFT can be decomposed into one-dimensional FrFT. This decomposition is crucial for \cono{}, as it relies on multidimensional FrFT (Thm. \ref{Theorem1}). (ii) Multiplication in the fractional domain is equivalent to convolution in the spatial domain for the proposed operator \cono{} (Thm. \ref{Theorem2}). (iii) The universal approximation capability for the \cono{} (Thm. \ref{Theorem3}). 

\begin{theorem}[Product Rule for FrFT] \label{Theorem1}
   Suppose $\mathcal{K}_\alpha(m_1, x_1, m_2, x_2 ..., m_n, x_n)$ denote the fractional integral kernel of the $N$-dimensional FrFT of a signal $f(x_1, x_2 ... x_n)$ with $m_1, m_2 ... m_n$ denoting the multidimensional FrFT coefficients, then the following fractional kernel property holds:
    \begin{equation}
        \mathcal{K}_\alpha(m_1, x_1, m_2, x_2 ..., m_n, x_n) = \prod_{i=1}^{N} \mathcal{K}_\alpha (m_i, x_i)
    \end{equation}
\end{theorem}
\textbf{Proof.} Refer to the App. \ref{proof: theorem1} for the proof. Thm. \ref{Theorem1}, show that an $N$-dimensional fractional integral kernel is a product of one-dimensional fractional integral kernels along each dimension. 

\begin{theorem}[Convolution Theorem for FrFT]\label{Theorem2}
    Suppose $f$ and $g$ are square-integrable functions. Define $e_{\alpha}(m) = e ^ {i \pi \mid m \mid ^2 cot(\alpha)}$ and $h(x) = (f * g)(x)$ where $*$ denotes the usual convolution integral with $\mathcal{F}^\alpha$, $\mathcal{G}^\alpha$, $\mathcal{H}^\alpha$ respectively denoting the FrFT of $f$, $g$ and $h$, respectively. Then,
    \begin{equation}
        \mathcal{H}^{\alpha}(m) = \mathcal{F}^{\alpha}(m) \mathcal{G}^{\alpha}(m) e_{-\alpha}(m). 
    \end{equation}
\end{theorem}

\textbf{Proof.} Refer to  App. \ref{proof: theorem2} for the proof. Thm. \ref{Theorem2} states that the convolution of two functions in the spatial domain is equivalent to the product of their respective fractional Fourier transforms, multiplied by a function dependent on the fractional order.

Finally, we present the universal approximation theorem of \cono{} below, similar to that for FNO \cite{kovachki2021universal}.

\begin{theorem}[Universal Approximation]\label{Theorem3}
    Let $s, s' > 0$ and $\alpha \in \mathbb{R}$; Suppose $\mathcal{G} : H^s(\mathcal{T}_{\alpha}^d;\mathbb{R}^{d_{\text{a}}}) \rightarrow H^{s'}(\mathcal{T}_{\alpha}^d;\mathbb{R}^{d_{\text{u}}})$ represent a continuous operator between Sobolev spaces where $\mathcal{T}_{\alpha}^d$ denotes the fractional order torus and $d_a, d_u \in \mathbb{N}$; and $K \subset H^s(\mathcal{T}_{\alpha}^d;\mathbb{R}^{d_{\text{a}}})$ is a compact subset. Then, for any $\varepsilon > 0$, there exists CoNO layers $\mathcal{N} : H^s(\mathcal{T}_{\alpha}^d;\mathbb{R}^{d_{\text{a}}}) \rightarrow H^{s'}(\mathcal{T}_{\alpha}^d;\mathbb{R}^{d_{\text{u}}})$ satisfying:
    \begin{equation}
        \sup_{v \in K} \|\mathcal{G}(v) - \mathcal{N}(v)\|_{L^2} \leq \varepsilon
    \end{equation}
\end{theorem}
\textbf{Proof.} Refer to App. \ref{proof: theorem3} for the proof. 
\vspace{-0.2cm}
\section{Numerical Experiments}
\vspace{-0.2cm}
This section provides a thorough empirical investigation of \cono{} in contrast to multiple vision models and neural operator baselines. We conduct extensive experiments on a diverse set of challenging benchmarks spanning various domains to demonstrate the efficacy of our proposed method.

\begin{table*}[t]
    \centering
    \caption{\label{tab:experiment_benchmarks} \small The main result with sixteen baselines on all benchmarks datasets: Mean Relative $\ell_2$ Error (Equation \ref{metric}) is reported as the evaluation metric, where a smaller $\ell_2$ Error indicates superior performance. "INCREMENT \%" refers to the relative error reduction concerning the second-best model on each benchmark. Specifically focusing on the 2D Navier–Stokes benchmark, a detailed comparison is conducted with KNO \cite{xiong2023koopman}, and TF-Net \cite{wang2020towards}, as they are designed for auto-regressive (time-dependent) tasks. Instances marked with '*' indicate that the baseline cannot handle the benchmark. In the color legend, \textcolor{blue}{blue} represents the best performance, \textcolor{green}{green} indicates the second-best performance, and \textcolor{orange}{orange} signifies the third-best performance among the baselines.}
    \resizebox{0.9\textwidth}{!}{
      \begin{tabular}{l|ccccccc}
        \toprule
        \multirow{1}{*}{\textbf{MODEL}} & \textbf{Elasticity-P} & \textbf{Elasticity-G} & \textbf{Plasticity} & \textbf{Navier-Stokes} & \textbf{Darcy} & \textbf{Airfoil} & \textbf{Pipe} \\
        \midrule
        \midrule
        U-NET \citeyearpar{ronneberger2015u} & 0.0235 & 0.0531 & 0.0051 & 0.1982 & 0.0080 & 0.0079 & 0.0065 \\
        RESNET \citeyearpar{he2016deep} & 0.0262 & 0.0843 & 0.0233 & 0.2753 & 0.0587 & 0.0391 & 0.0120 \\
        TF-NET \citeyearpar{wang2020towards} & \textbackslash & \textbackslash & \textbackslash & 0.1801 & \textbackslash & \textbackslash & \textbackslash \\
        SWIN  \citeyearpar{liu2021swin} & 0.0283 & 0.0819 & 0.0170 & 0.2248 & 0.0397 & 0.0270 & 0.0109 \\
        DEEPONET \citeyearpar{lu2021learning} & 0.0965 & 0.0900 & 0.0135 & 0.2972 & 0.0588 & 0.0385 & 0.0097 \\
        FNO \citeyearpar{li2020fourier} & \cellcolor[HTML]{FFE5CC} 0.0229 & 0.0508 & 0.0074 & 0.1556 & 0.0108 & 0.0138 & 0.0067 \\
        U-FNO \citeyearpar{wen2022u} & 0.0239 & 0.0480 & 0.0039 & 0.2231 & 0.0183 & 0.0269 & \cellcolor[HTML]{FFE5CC} 0.0056 \\
        WMT \citeyearpar{gupta2021multiwavelet} & 0.0359 & 0.0520 & 0.0076 & \cellcolor[HTML]{FFE5CC} 0.1541 & 0.0082 & 0.0075 & 0.0077 \\
        GALERKIN \citeyearpar{cao2021choose} & 0.0240 & 0.1681 & 0.0120 & 0.2684 & 0.0170 & 0.0118 & 0.0098 \\
        SNO \citeyearpar{fanaskov2022spectral} & 0.0390 & 0.0987 & 0.0070 & 0.2568 & 0.0495 & 0.0893 & 0.0294 \\
        U-NO \citeyearpar{rahman2022u} & 0.0258 & \cellcolor[HTML]{FFE5CC} 0.0469 & \cellcolor[HTML]{FFE5CC} 0.0034 & 0.1713 & 0.0113 & 0.0078 & 0.0100 \\
        HT-NET \citeyearpar{liu2022ht} & 0.0372 & 0.0472 & 0.0333 & 0.1847 & 0.0079 & \cellcolor[HTML]{FFE5CC}0.0065 & 0.0059 \\
        F-FNO \citeyearpar{tran2021factorized} & 0.0263 & 0.0475 & 0.0047 & 0.2322 & \cellcolor[HTML]{FFE5CC} 0.0077 & 0.0078 & 0.0070 \\
        KNO \citeyearpar{xiong2023koopman} & \textbackslash & \textbackslash & \textbackslash & 0.2023 & \textbackslash & \textbackslash & \textbackslash \\
        GNOT \citeyearpar{hao2023gnot} & 0.0315 & 0.0494 & * & 0.1670 & 0.0105 & 0.0081 & *\\
        LSM \citeyearpar{wu2023solving} & \cellcolor[HTML]{E5FFD9} 0.0218 & \cellcolor[HTML]{DDEEFF} \textbf{0.0408} & \cellcolor[HTML]{E5FFD9} 0.0025 & \cellcolor[HTML]{E5FFD9} 0.1535 & \cellcolor[HTML]{E5FFD9} 0.0065 & \cellcolor[HTML]{E5FFD9} 0.0062 & \cellcolor[HTML]{DDEEFF} \textbf{0.0050} \\
        \textbf{\cono{} (Ours)} &  \cellcolor[HTML]{DDEEFF} \textbf{0.0210} & \cellcolor[HTML]{E5FFD9} 0.0436 & \cellcolor[HTML]{DDEEFF} \textbf{0.0019} &  \cellcolor[HTML]{DDEEFF} \textbf{0.1287} &  \cellcolor[HTML]{DDEEFF} \textbf{0.0051} & \cellcolor[HTML]{DDEEFF} \textbf{0.0057} & \cellcolor[HTML]{E5FFD9} 0.0054\\
        \midrule
        \midrule
         \textbf{INCREMENT} \% & \textbf{3.8\%} & -6.8\% & \textbf{31.6\%} & \textbf{19.3\%} & \textbf{27.5\%} & \textbf{8.7\%} & -8.0\% \\
        \bottomrule
  \end{tabular}}
\end{table*}

\vspace{-0.2cm}
\subsection{Experiments Details and Main Result}
\vspace{-0.1cm}
\textbf{Benchmarks:} We assess the performance of our model on Darcy and Navier Stokes \cite{li2020fourier} benchmarks to gauge its proficiency on regular grids. Subsequently, we extend our experimentation to benchmarks featuring irregular geometries, such as Airfoil, Plasticity, and Pipe \cite{li2022fourier}, modeled using structured meshes and Elasticity \cite{li2022fourier}, represented in point clouds. Refer to App. section \ref{Appendix C} for more details about benchmarks and tasks.

\textbf{Baselines:} We assess \cono{} by comparing it against sixteen established models across seven benchmarks, which include baselines from vision models (U-Net \cite{ronneberger2015u}, ResNet \cite{he2016deep}, SwinTransformer \cite{liu2021swin}) and thirteen baselines specifically designed for PDEs (DeepONet \cite{lu2021learning}, TF-Net \cite{wang2020towards}, FNO \cite{li2020fourier}, U-FNO \cite{wen2022u}, WMT \cite{gupta2021multiwavelet}, GalerkinTransformer \cite{cao2021choose}, SNO \cite{fanaskov2022spectral}, U-NO \cite{rahman2022u}, HT-Net \cite{liu2022ht}, F-FNO \cite{tran2021factorized}, KNO \cite{xiong2023koopman}, GNOT \cite{hao2023gnot}, LSM \cite{wu2023solving}). Notably, for the Elasticity-P benchmark in the point cloud, we incorporate the specialized transformation proposed by geo-FNO \cite{li2022fourier} at both the start and end of these models. This transformation facilitates the conversion of irregular input domains into or back from a uniform mesh.

\textbf{Evaluation Metric:} Mean relative $\ell_2$ error is used throughout the experiments.
\begin{equation}
    \mathcal{L}  = \frac{1}{N} \sum_{i=1}^{N} \frac{\| \mathcal{G}_\theta(a_i) -  \mathcal{G}^\dagger(a_i)\|_2}{\| \mathcal{G}^\dagger(a_i) \|_2} \label{metric}
\end{equation}
the regular mean-squared error (MSE) is enhanced with a normalizer $\|\mathcal{G}^\dagger(a_i)\|_2$ to take account for discrepancies in absolute resolution scale across different benchmarks as described in \cite{li2020fourier}.

\textbf{Implementation Details:} We have used mean relative $\ell_2$ error (Eq. \ref{metric}) as the training and evaluation metric. We train all the models for 500 epochs using the Adam optimizer \cite{kingma2014adam}. Comprehensive details are provided in the App. section \ref{Appendix D}. All the experiments are conducted on a Linux machine running Ubuntu 20.04.3 LTS on an Intel(R) Core(TM) i9-10900X processor and a single NVIDIA RTX A6000 GPU with 48 GB RAM. 

\textbf{Empirical Results:} As illustrated in Table \ref{tab:experiment_benchmarks}, \cono{} consistently demonstrates superior performance on all the datasets in comparison to all baseline models, ranking among the top two. This performance superiority is evident across benchmark datasets characterized by diverse geometries and dimensions, exhibiting an average improvement of 10.9\%. With the second-best performance observed in the Pipe and Elasticity-G benchmarks, our findings suggest that architectures resembling UNET demonstrate superior efficacy in capturing the underlying solution compared to the \cono{} model. When applied to time-dependent PDEs, \cono{} surpasses all previously established baselines, achieving an average improvement of 25.5\%. This result underscores the efficacy of incorporating the change in frequency derivative captured by the FrFT in complex domains, thereby showcasing the promise of our approach in handling temporal dynamics in PDEs.

\vspace{-0.2cm}
\subsection{Ablation Study and Additional Results}
\vspace{-0.1cm}
\begin{wraptable}{r}{7cm}
    \centering
    \vspace{-0.7cm}
    \caption{\label{tab:ablation} \small Comprehensive ablation study on \cono{}: investigating the impact of individual component removal on Navier-Stokes and Darcy Flow benchmark (w/o denotes the performance without that component).}
    \resizebox{0.4\textwidth}{!}{
    \begin{tabular}{l|cc}
        \toprule
        \textbf{DESIGN} & \textbf{Navier-Stokes} & \textbf{Darcy} \\
        \midrule
        \midrule
        w/o Bias  & 0.1425 & 0.0080 \\
        w/o FrFT                  & 0.1535 & 0.0086 \\
        w/o Complex NN            & 0.1390 & 0.0072 \\
        w/o Alias Free Activation  & 0.1295 & 0.0052 \\
        \midrule
        \midrule
        \textbf{\cono{}}           & 0.1287 & 0.0050 \\
        \bottomrule
    \end{tabular}}
\end{wraptable}

\begin{figure*}[t]
    \centering
    \includegraphics[width=0.9\linewidth]{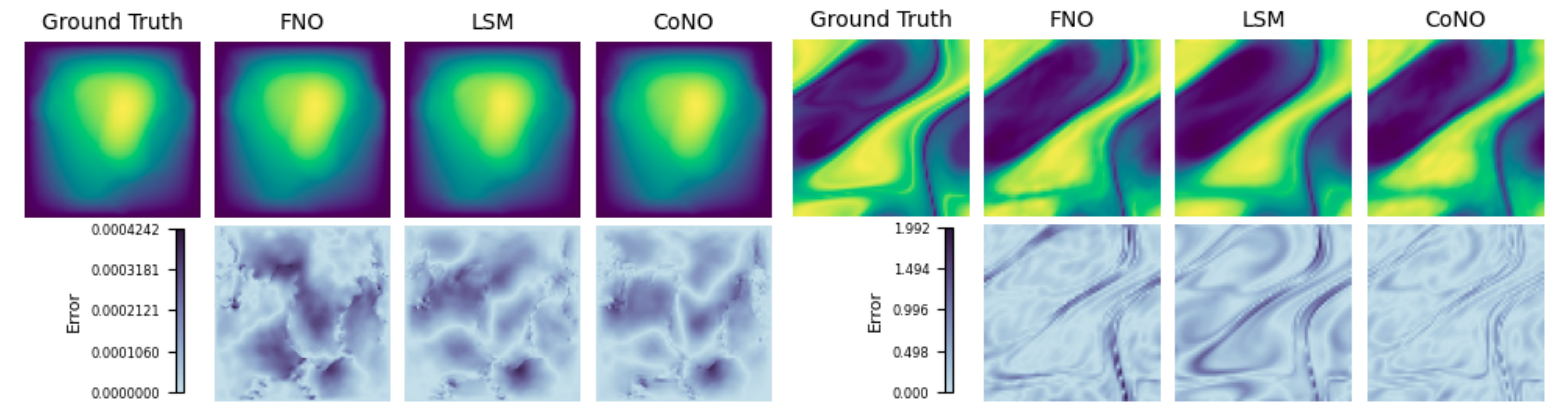}
    \caption{\label{fig:figure3} Depiction of results for different methods on fluid datasets, Darcy Flow \textbf{(Left)} and Navier Stokes \textbf{(Right)}. We plotted the heatmap of the absolute difference value between ground truth and prediction to compare the predicted output. See the App. Fig. \ref{fig:figure10} for more solid physics and fluid physics benchmarks showcases.}
    \vspace{-0.5cm}
\end{figure*}

To assess the efficacy of each component in the \cono{} operator, we conducted a comprehensive ablation study by systematically excluding individual components. The results presented in Table ~\ref{tab:ablation} indicate that all components are crucial for the effectiveness of the \cono{} operator, as evidenced by a notable change in the $\ell_2$ error with the addition or deletion of an element. Specifically, removing the FrFT block results in a substantial degradation in performance, underscoring its effectiveness in capturing non-stationary signals. Similarly, the absence of the CVNN leads to comparable adverse effects. Notably, our analysis reveals that including bias is instrumental in introducing high-frequency components, further emphasizing its importance. Interestingly, \cono{} performance degraded only slightly after removing the alias-free activation function as defined. It raises an intriguing question regarding its necessity for optimizing the operator's efficiency.

\textbf{Visual Demonstrations:} For a concise representation of intuitive performance, Fig. \ref{fig:figure3} presents a comparative analysis between FNO, LSM, and \cono{}. Notably, \cono{} exhibits remarkable proficiency in addressing time-dependent PDEs, including Navier-Stokes and Plasticity, as depicted in App. Fig. \ref{fig:figure10}. Moreover, \cono{} outperforms LSM and FNO significantly in the case of Darcy by 27\% and having fewer artifacts present in prediction. Additionally, \cono{} excels in capturing singularities around corners, as illustrated in the elasticity dataset in the App. Fig. \ref{fig:figure10}, emphasizing its robust and superior performance.

\begin{figure*}[t]
    \centering
    \includegraphics[width=1.0\linewidth]{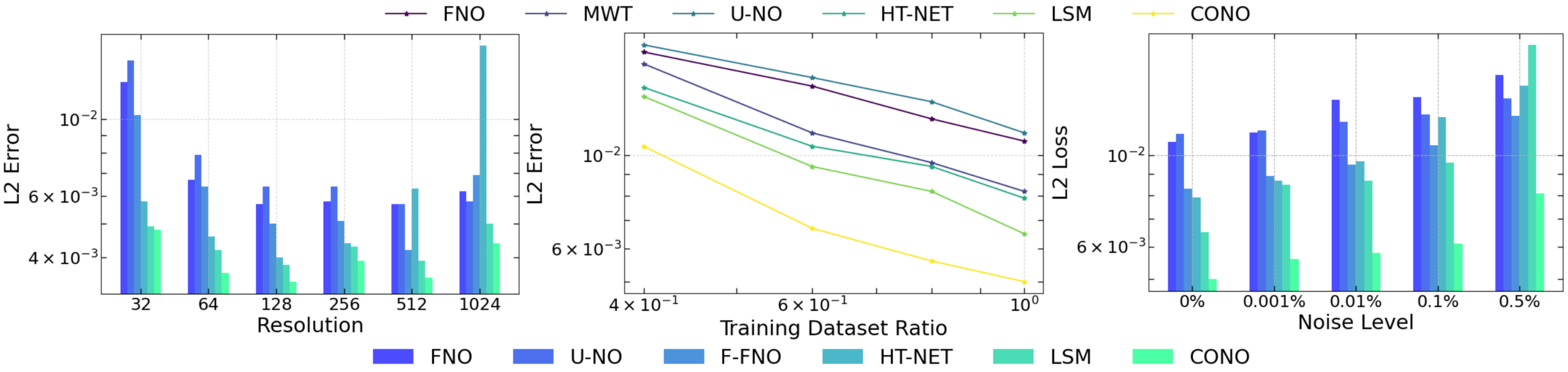}
    \vspace{-0.7cm}
    \caption{\label{fig:mainplot} \small \textbf{(Left)} Models performance under different resolutions on Darcy. \textbf{(Middle)} Models performance under different training datasets the ratio on Darcy. \textbf{(Right)} Models performance under the presence of noise on Darcy. The lower $l2$ loss indicates better performance.}
    \vspace{-0.6cm}
\end{figure*}

\textbf{Performance across various Resolutions:} 
In Fig. \ref{fig:mainplot} (Left), the operator \cono{} consistently exhibits superior performance compared to other operators on darcy flow PDEs at various resolutions. Notably, \cono{} demonstrates stability across different resolutions, adhering to the principle of discrete-continuous equivalence as shown in App. section \ref{sec:resolution}. It contrasts HT-Net, which experiences degradation in very high dimensions. Furthermore, FNO and \cono{} represent the exclusive class of operators capable of zero-shot super-resolution without requiring explicit training.

\textbf{Out of Distribution Generalization:} We conducted experiments on the Navier-Stokes dataset in this investigation, training our model with a viscosity coefficient of $10^{-5}$. Subsequently, we assessed the out-of-distribution generalization capabilities by evaluating the trained model on a viscosity coefficient of $10^{-4}$. Our findings consistently reveal that \cono{} demonstrates a significantly superior generalization performance, exhibiting an increment of $64.3\%$ compared to FNO. It also highlights the significance of capturing latent variable information or the UNET architecture, as achieved by LSM, which outperforms all other operators even \cono{} as shown in App. section \ref{sec:out}.

\textbf{Data Efficiency:} As demonstrated in Fig. \ref{fig:mainplot} (Middle), \cono{} exhibits comparable performance to the second-best operator LSM when trained on 60\% of the data. Furthermore, across various training dataset ratios, \cono{} consistently outperforms all other operators, underscoring its superior data efficiency compared to SOTA operators as demonstrated in App. section \ref{sec:data}.

\textbf{Robustness to Noise:} In this study, we performed experiments introducing different noise levels into the training using Gaussian noise. The noise addition process follows: For each input sample denoted as $x(n)$ within the dataset $D$, we modified it by adding Gaussian noise with parameters $\gamma N(0, \sigma^2_D)$. Here, $\sigma^2_D$ represents the variance of the entire dataset, and $\gamma$ indicates the specified noise intensity level. Our investigation yielded notable results as in Fig. \ref{fig:mainplot} (Right), particularly when evaluating the performance of \cono{} in the presence of noise within the training dataset; specifically, the noisy training with 0.1\% yielded a better result than the LSM operator without noisy training, confirming the robustness of \cono{} to the noisy dataset shown in App. section \ref{sec:noise}.

\begin{wrapfigure}{l}{0.5\textwidth}
    \centering
    \includegraphics[width=1.0\linewidth]{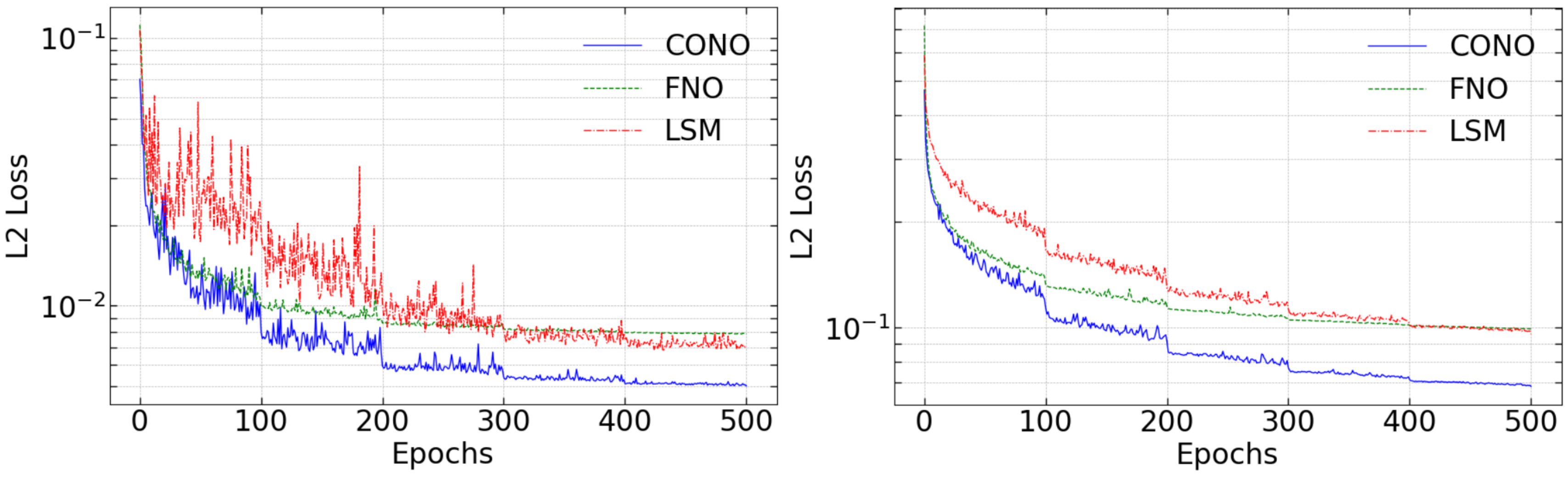}
    \vspace{-0.7cm}
    \caption{\label{fig:training_curve} \small Learning Curve for Darcy flow \textbf{(Left)} and Navier Stokes \textbf{(Right)} where the x-axis denotes epochs and the y-axis $l2$ error.}
\end{wrapfigure}

\textbf{Training Stability:} The performance of \cono{} is more stable during training, as visually observed in Fig. \ref{fig:training_curve}. The model exhibits reduced oscillations and consistently performs better than FNO and LSM. Remarkably, \cono{} attains an equivalent performance compared with the best LSM performance after training within the initial 200 epochs, demonstrating its efficient faster and better convergence while training.

\textbf{Long Time Prediction on Navier Stokes:} 
We evaluated the long-term behavior of the proposed operator \cono{} by training it on the Navier-Stokes equation (viscosity coefficient of $10^{-4}$), as shown in the App. section \ref{sec:long} shows that \cono{} excels in extrapolating beyond the prediction horizon compared to LSM and FNO.
\vspace{-0.3cm}
\section{Conclusion}
\vspace{-0.3cm}
Altogether, we introduce a new operator learning paradigm called the \cono{}, which capitalizes on CVNNs and the FrFT as the integral operator. We theoretically prove that \cono{} follows the universal approximation theorem. We demonstrate the effectiveness of leveraging the expressive power of CVNNs within the operator learning framework to construct resilient, data-efficient, and superior neural operators capable of improving the learning of function-to-function mappings. Empirically, we show that \cono{} presents the SOTA results in terms of performance, zero-shot super-resolution, out-of-distribution generalization, and noise robustness. 
This advancement makes \cono{} a promising method for developing efficient operators for real-time PDE inference, offering new tools for the SciML community.

\textbf{Limitations and future work.} Although \cono{} shows improved empirical performance, the specific features that enable this superiority remain unclear. Understanding the loss landscape and learning mechanisms behind this performance is crucial. Additionally, making \cono{} computationally efficient is essential to accelerate inference. Additional future work and limitations are discussed in App. section \ref{Appendix G}. Further, the broader impacts of the work are discussed in App. section \ref{Appendix H}.

\bibliographystyle{plainnat}
\small{\bibliography{references}}

\newpage
\appendix

\section*{Appendix} \label{Appendix}

\section{Fractional Fourier Transform} \label{Appendix A}
\begin{figure*}[!htb]
    \centering
    \includegraphics[width=1.0\linewidth]{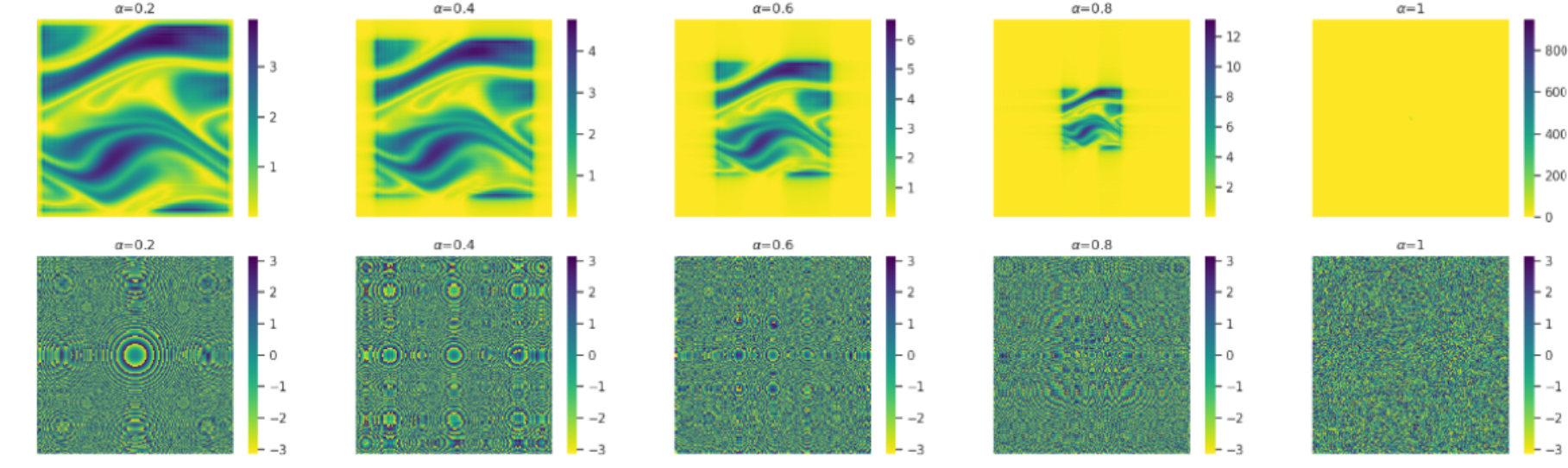}
    \caption{\label{fig:figure6} \small The figure illustrates the FrFT of the Navier-Stokes Partial Differential Equation (PDE) across varying $\alpha$ values. The \textbf{(Top)} displays heatmaps representing the magnitude of the transformed data, while the \textbf{(Bottom)} showcases corresponding heatmaps depicting the phase angle. Each column corresponds to a distinct $\alpha$ value, highlighting the impact of fractional Fourier transform parameters on the learned representation of the Navier-Stokes PDE. It is observed that in FT, most spectrum is concentrated at low frequency for $\alpha = 1$.}
\end{figure*}

\begin{wrapfigure}{l}{0.5\textwidth}
    \centering
    \includegraphics[width=0.8\linewidth]{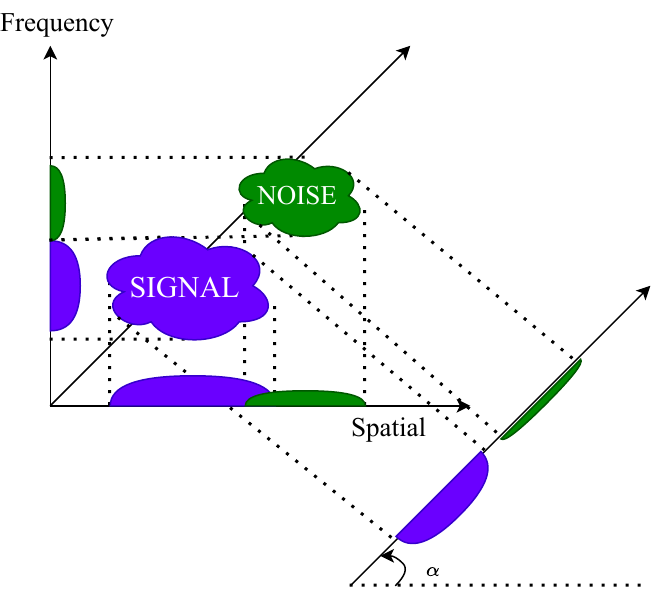}
    \caption{\label{fig:figure7} \small The figure illustrates the inherent optimality of FrFT in signal filtering for non-stationary image signals, showcasing its ability to separate effects along the fractional order axis and examining its relationship with spatial-frequency plane rotation.}
\end{wrapfigure}

Fig. \ref{fig:figure7} is an elucidative representation of the intrinsic optimality associated with the Fractional Fourier Transform (FrFT) in signal filtering for non-stationary image signals. It notably underscores FrFT's distinctive capability to separate various signal effects along the fractional order axis effectively. Furthermore, the figure provides insightful scrutiny into the intricate relationship between FrFT and the rotation of the spatial-frequency plane. This visualization contributes to a nuanced understanding of FrFT's prowess in handling non-stationary image signals. It offers valuable insights into its utility and potential applications in the broader signal-processing domain. Fig. \ref{fig:figure6} shows that as the parameter $\alpha$ increases, the spectral distribution becomes increasingly concentrated around the origin. This concentration results in a loss of spatial information while emphasizing frequency information. Conversely, lower values of $\alpha$ reveal the preservation of information in the phase domain, which, unfortunately, becomes increasingly noisy when transformed into the frequency domain. This behavior highlights the delicate balance between spatial and frequency information, emphasizing the parameter $\alpha$'s critical role in shaping the signal's characteristics.

\section{Mathematical Analysis} \label{Appendix B}
In this section, we will first present the necessary definitions and then prove the theorem as stated in the main content of the paper. Subsequently, we present a theoretical analysis of several components of the proposed method. 

\begin{definition}[Fractional Torus]\cite{fu2022convergence}
    Let consider $\alpha \in \mathbb{R}$, and $\alpha \neq n 	\mathbb{Z}$. Define fractional torus of order $\alpha$, denoted as $\mathcal{T}_{\alpha} ^ {n}$ is the cube $ [ 0, \mid \sin{\alpha} \mid ]^{n}$ with opposite sides identified. We will define FrFT on $\mathcal{T}_{\alpha} ^{n}$ for the rest of the analysis. We will define Fractional convolution and fractional approximation in $L^{p} (\mathcal{T}_{\alpha} ^{n})$ for $ 1 \leq p \leq \infty$.
\end{definition} 

\begin{definition}
    Let $\alpha \in \mathcal{R}$ and $\alpha \neq n$. Define $e_{\alpha}(x) = e ^ {i \pi \mid x \mid ^2 cot \alpha}$ and $e_{\alpha} f(x) = e_{\alpha}(x) f(x)$ for function on $\mathcal{T}_{\alpha}^{n}$.
\end{definition}

\begin{definition}
    Let $1 \leq p \leq \infty$. A function $f$ on $\mathcal{T}_{\alpha} ^{n}$ lies in the space $e_{-\alpha} L^{p} (\mathcal{T}_{\alpha} ^{n})$ if 
    \begin{equation}
        f(x) = e_{-\alpha} g(x), \: \: g \in L^{p} (\mathcal{T}_{\alpha} ^{n})
    \end{equation}
    and $\mid \mid f \mid \mid _{L^{p} (\mathcal{T}_{\alpha} ^{n})} < \infty$.
\end{definition}

\begin{definition}[Fractional Fourier Transform on Fractional Torus \citet{fu2022convergence}]
    For complex-valued functions $f \in e_{-\alpha} L^{1} (\mathcal{T}_{\alpha} ^{n})$, $\alpha \in \mathbb{R}$ and $m \in \mathbb{Z}^{n}$, we define,
    
    \begin{equation}
        \mathcal{F}^{\alpha} (f)(m) = 
        \begin{cases}
            \int_{\mathcal{T}_{\alpha} ^{n}} f(x) K_\alpha(m, x) dx &  \alpha \neq \pi \mathbb(Z) \\
            f(m) &  \alpha = 2 \pi \mathbb(Z)\\
            f(-m) &  \alpha = 2 \pi \mathbb(Z) + \pi
        \end{cases} \label{eq: frft torus}
    \end{equation}

    where, 
    \begin{equation}
        K_\alpha(m, x) = A_{\alpha}^{n} e_{\alpha}(x) e_{\alpha} (m, x) e_{\alpha}(m)
    \end{equation}

    here, $A_{\alpha} = \sqrt{1 - i \cot{\alpha}}$ and $e_{\alpha} (m, x) = e ^{-i 2\pi (m.x) \csc{\alpha}}$. where $\mathcal{F}^{\alpha} (f)(m)$ denotes the $m^{th}$ fractional fourier coefficient of order $\alpha$ for $f \in e_{-\alpha} L^{1} (\mathcal{T}_{\alpha} ^{n})$. 
\end{definition}

\begin{remark}
    For Eq. \ref{eq: frft torus}, it reduces to \textbf{Fourier Transform (FT)} when $\alpha = \frac{\pi}{2}$, $\cot{\alpha} = 0$ and $\csc{\alpha} = 1$, where
    \begin{equation}
        K_{\alpha}(m, x) = e ^{-i 2\pi (m.x)}
    \end{equation}
\end{remark}

\begin{proposition}[Linear and Bounded Operator]
Let $f$, $g$ be in $e_{-\alpha} L^{1} (\mathcal{T}_{\alpha} ^{n})$. Then for $m, k \in \mathbb{Z}^{n}$, $\lambda \in \mathbb{C}$, $y \in \mathcal{T}_{\alpha} ^{n}$ then following holds true.
\begin{enumerate}
    \item $\mathcal{F}^{\alpha} (f + g)(m)$ = $\mathcal{F}^{\alpha} (f)(m) + \mathcal{F}^{\alpha} (g)(m)$.
    \item $\mathcal{F}^{\alpha} (\lambda f)(m) = \lambda \mathcal{F}^{\alpha} (f)(m)$.
    \item $\sup_{m \in \mathbb{Z}^{n}} | \mathcal{F}^{\alpha} (f)(m) | \leq | \csc{\alpha} | ^ {\frac{n}{2}} \mid \mid f \mid \mid _{L^{1} (\mathcal{T}_{\alpha} ^{n})}$
\end{enumerate}
\end{proposition}

\textbf{Proof:} For proof refer \citet{fu2022convergence}.

\begin{proposition}[Uniqueness of Fractional Fourier Transform]
    If $f, g \in e_{-\alpha}L^{1}(\mathcal{T}_{\alpha}^{n})$ satisfy $\mathcal{F}^{\alpha}(f)(m) = \mathcal{F}^{\alpha}(g)(m)$ for all $m \in \mathbb{Z}^{n}$, then $f = g$ a.e.
\end{proposition}
\textbf{Proof:} For proof refer \citet{fu2022convergence}.

\begin{corollary}[Fractional Fourier Inversion]
    If $f \in e_{-\alpha}L^{1}(\mathcal{T}_{\alpha}^{n})$ and 
    \begin{equation}
        \sum_{m \in \mathbb{Z}^{n}} | \mathcal{F}^{\alpha}(f)(m) | < \infty.
    \end{equation}
    Then,
    \begin{equation}
        f(x) = \sum_{m \in \mathbb{Z}^{n}} \mathcal{F}^{\alpha}(f)(m) K_{-\alpha}(m, x) \: \text{a.e.}
    \end{equation}
    and therefore, $f$ is almost everywhere, equal to a continuous function.
\end{corollary}

\begin{theorem}[\textbf{Product Rule}] \label{proof: theorem1}
    Suppose $\mathcal{K}_\alpha(m_1, x_1, m_2, x_2 ..., m_n, x_n)$ denote the fractional integral kernel of the $N$-dimensional FrFT of a signal $f(x_1, x_2 ... x_n)$ with $m_1, m_2 ... m_n$ denoting the multidimensional FrFT coefficients, then the following fractional kernel property holds:
    \begin{equation}
        \mathcal{K}_\alpha(m_1, x_1, m_2, x_2 ..., m_n, x_n) = \prod_{i=1}^{N} \mathcal{K}_\alpha (m_i, x_i)
    \end{equation}
\end{theorem}
\textbf{Proof: }

Let consider the $N$-dimensional signal $f(x_1, x_2, ..., x_n)$,

Then, let's define the FrFT of the signal as follows:

\begin{equation}
    \mathcal{F}^{\alpha} (f)(m_1, m_2, ..., m_N) = \int_{-\infty} ^ {\infty} ... \int_{-\infty}^{\infty} f(x_1, x_2, ..., x_N) \prod_{i=1}^{N} \mathcal{K}_\alpha (m_i, x_i) dx_i, 
    \label{equ:19}
\end{equation}

From Eq. \ref{equ:19}, we can easily prove for $\alpha = 0$,

\begin{equation}
    \mathcal{F}^{\alpha} (f) (m_1, m_2, ..., m_N) = f(x_1, x_2, ..., x_n)
\end{equation}

and for $\alpha=1$, it reduces to \textit{Fourier Transform (FT)}.

Therefore, if we can show that $\mathcal{F}^{\alpha}. \mathcal{F}^{\beta} = \mathcal{F}^{\alpha + \beta}$ then obviously $\mathcal{F}^{\alpha}$ satisfy all the FrFT properties.

Then,
\begin{equation}
    \mathcal{F}^{\alpha}. \mathcal{F}^{\beta} (f) (m_1, m_2, ..., m_n) = \int_{-\infty} ^ {\infty} ... \int_{-\infty} ^ {\infty} \prod_{i=1}^{N} \mathcal{K}_\alpha (m_i, x_i) \int_{-\infty} ^ {\infty} ... \int_{-\infty}^{\infty} f(y_1, y_2, ..., y_N) \prod_{i=1}^{N} \mathcal{K}_\beta (x_i, y_i) dy_i dx_i 
\end{equation}

\begin{equation}
    \mathcal{F}^{\alpha}. \mathcal{F}^{\beta} (f) (m_1, m_2, ..., m_n) =  \int_{-\infty} ^ {\infty} ... \int_{-\infty} ^ {\infty} f(y_1, y_2, ..., y_N) \{ \prod_{i=1}^{N} \int_{-\infty} ^ {\infty} \mathcal{K}_\alpha (m_i, x_i) \mathcal{K}_\beta (x_i, y_i) dx_i \} dy_1 dy_2 ...dy_n
\end{equation}

Now using the result from \cite{almeida1993introduction}, 

\begin{equation}
    \int_{-\infty} ^ {\infty} \mathcal{K}_\alpha (m, x) K_\beta (x, y) du = \mathcal{K}_{\alpha + \beta}(m, y)
\end{equation}

Using the above property, 

\begin{equation}
    \mathcal{F}^{\alpha}. \mathcal{F}^{\beta} (f) (m_1, m_2, ..., m_n) =  \int_{-\infty} ^ {\infty} ... \int_{-\infty} ^ {\infty} f(y_1, y_2, ..., y_N) \prod_{i=1}^{N} K_{\alpha + \beta}(m_i, y_i) dy_1 dy_2 ...dy_n = \mathcal{F}^{\alpha + \beta}
\end{equation}

Therefore, we conclude that,
\begin{equation}
        \mathcal{K}_\alpha(m_1, x_1, m_2, x_2 ..., m_n, x_n) = \prod_{i=1}^{N} \mathcal{K}_\alpha (m_i, x_i)
\end{equation}

\begin{definition}
Let $f$ and $g$ be square integrable functions, Let,  $h(x) = (f * g)(x)$ (where $*$ denotes the convolution), i.e., 
\begin{equation}
    h(x) = (f * g)(x) = \int_{-\infty}^{\infty} f(t)g(x - t) \, dt. \label{equ: convolution}
\end{equation}
\end{definition}

\begin{definition}
For any function $f(x)$, let us define the function $e_{\alpha} f(x) = e_{\alpha}(x) f(x)$. For any two functions $f$ and $g$, we define the convolution operation $\ast$ by
\begin{equation}
    h(x) = (f \ast g)(x) = A_{\alpha} e_{-\alpha}(x) (e_{\alpha} f \ast e_{\alpha} g) (x),
\end{equation}
where $\ast$ is the convolution operation as defined in Eq. \ref{equ: convolution}.
\end{definition}

\begin{theorem}[\textbf{Convolution Theorem for Fractional Fourier Transform (FrFT)}] \label{proof: theorem2}
    Suppose $f$ and $g$ are square-integrable functions. Define $e_{\alpha}(m) = e ^ {i \pi \mid m \mid ^2 cot(\alpha)}$ and $h(x) = (f * g)(x)$ where $*$ denotes the usual convolution integral with $\mathcal{F}^\alpha$, $\mathcal{G}^\alpha$, $\mathcal{H}^\alpha$ respectively denoting the FrFT of $f$, $g$ and $h$, respectively. Then,
    \begin{equation}
        \mathcal{H}^{\alpha}(m) = \mathcal{F}^{\alpha}(m) \mathcal{G}^{\alpha}(m) e_{-\alpha}(m). 
    \end{equation}
\end{theorem}

\textbf{Proof:} By FrFT definition from Eq. \ref{eq:frft}, we know that

\begin{equation}
    \begin{split}
        \mathcal{H}^{\alpha}(h)(m) &= A_{\alpha} \int_{-\infty}^{\infty} h(t) e_{\alpha}(t) e_{\alpha} (m, t) e_{\alpha}(m) dt \\
        &= A_{\alpha}^2 \int_{-\infty}^{\infty} e_{\alpha}(t) e_{\alpha} (m, t) e_{\alpha}(m) e_{-\alpha}(t) dt \int_{-\infty}^{\infty} e_{\alpha}(x) f(x) e_{\alpha}(t-x) g(t-x) dx \\
        &= A_{\alpha}^2 \int_{-\infty}^{\infty} \int_{-\infty}^{\infty} f(x) g(t-x)  e_{\alpha}(t) e_{\alpha} (m, t) e_{\alpha}(m) e_{-\alpha}(t) e_{\alpha}(x) e_{\alpha}(t-x) dx dt
    \end{split}
\end{equation}

By substituting $t - x = v$ in the above eq. , we obtain

\begin{equation}
    \begin{split}
        \mathcal{H}^{\alpha}(h)(m) &= A_{\alpha}^2 e_{\alpha}(m) \int_{-\infty}^{\infty} \int_{-\infty}^{\infty} f(x) g(v) e_{\alpha}(x + v) e_{\alpha} (m, x + v) e_{-\alpha}(x + v) e_{\alpha}(x) e_{\alpha}(v) dx dv\\
        &= A_{\alpha}^2 e_{-\alpha}(m) \int_{-\infty}^{\infty} f(x)  e_{\alpha}(x) e_{\alpha}(m) e_{\alpha} (m, x) dx \int_{-\infty}^{\infty} g(v) e_{\alpha}(v) e_{\alpha}(m) e_{\alpha} (m, v) dv \\
        &= \mathcal{F}^{\alpha}(m) \mathcal{G}^{\alpha}(m) e_{-\alpha}(m)
    \end{split}
\end{equation}
Therefore, we have, 

\begin{equation}
    \mathcal{H}^{\alpha}(m) = \mathcal{F}^{\alpha}(m) \mathcal{G}^{\alpha}(m) e_{-\alpha}(m).
\end{equation}

\begin{definition}[Fractional Fourier Projection Operator]
    Let's define the Fractional Fourier wave-number as follows:
    \begin{equation}
        \mathcal{K}_{N} = \{ k \in \mathbb{Z}^{n}: | k |_{\infty} \le N \}
    \end{equation}
    And define fractional projection operator to truncate high order coefficient of a function as follows:
    \begin{equation}
        \mathcal{P}_{N}(f)(x) = \mathcal{F}^{-\alpha}( \mathcal{F}^{\alpha}(f)(k). \boldsymbol{1}_{\mathcal{K}_{N}} (k))
    \end{equation}
    where $1_{\mathcal{K}_{N}} (k)$ indicates indicator function which takes 1 when $k \in \mathcal{K}_{N}$ and $0$ otherwise.
\end{definition}

\begin{theorem}[Convergence of Fractional Fourier Series]\label{convergence}
    Let $\alpha \in \mathbb{R}$ and $\alpha \neq \pi \mathbb{Z}$, and $f \in e_{-\alpha}L^{p}(\mathcal{T}_{\alpha}^{n})$ ($1 \leq p < \infty$), $\mathcal{P}_{N}(f)(x)$ denotes as follows:
    \begin{equation}
        \mathcal{P}_{N}(f)(x) = \mathcal{F}^{-\alpha}( \mathcal{F}^{\alpha}(f)(k). \boldsymbol{1}_{\mathcal{K}_{N}} (k))
    \end{equation}
    where, $\mathcal{F}^{\alpha}(f)(k)$ denotes the $k^{th}$ fractional coefficient. Then,
    \begin{equation}
        \lim_{N \rightarrow \infty} \sup_{x} ||f(x) - \mathcal{P}_{N}(f)(x)|| \rightarrow 0
    \end{equation}
\end{theorem}

\begin{theorem}[\textbf{Universal Approximation of CoNO}]\label{proof: theorem3}
    Let $s, s' > 0$ and $\alpha \in \mathbb{R}$; Suppose $\mathcal{G} : H^s(\mathcal{T}_{\alpha}^d;\mathbb{R}^{d_{\text{a}}}) \rightarrow H^{s'}(\mathcal{T}_{\alpha}^d;\mathbb{R}^{d_{\text{u}}})$ represent a continuous operator between Sobolev spaces where $\mathcal{T}_{\alpha}^d$ denotes the fractional order torus and $d_a, d_u \in \mathbb{N}$; and $K \subset H^s(\mathcal{T}_{\alpha}^d;\mathbb{R}^{d_{\text{in}}})$ is a compact subset. Then, for any $\varepsilon > 0$, there exists CoNO layers $\mathcal{N} : H^s(\mathcal{T}_{\alpha}^d;\mathbb{R}^{d_{\text{in}}}) \rightarrow H^{s'}(\mathcal{T}_{\alpha}^d;\mathbb{R}^{d_{\text{out}}})$ satisfying:
    \begin{equation}
        \sup_{v \in K} \|\mathcal{G}(v) - \mathcal{N}(v)\|_{L^2} \leq \varepsilon
    \end{equation}
\end{theorem}

\textbf{Proof:} Let $\alpha \in \mathbb{R}$ and define the following fractional orthogonal projection operator:
\begin{equation}
    \mathcal{G}_{N}: H^s(\mathcal{T}_{\alpha}^d) \rightarrow H^{s'}(\mathcal{T}_{\alpha}^d), \: \: \mathcal{G}_{N}(v) = \mathcal{P}_{N}\mathcal{G}(\mathcal{P}_{N}(v))
\end{equation}

Using Thm. \ref{convergence}, we have for $\forall \varepsilon$, there exists $N \geq 0$ such that:
\begin{equation}
    \|\mathcal{G}(v) - \mathcal{G}_{N}(v)\|_{L^2} \leq \varepsilon,  \:\: \forall v \in K
\end{equation}

We need to find $\mathcal{N}$ to approximate $\mathcal{G}_{N}$.

Next we define \textit{Fractional Conjugate} of $\mathcal{G}_{N}$ as follows:

\begin{equation}
    \mathcal{\hat{G}}_{N}: \mathbb{C}^{\mathcal{K}_{N}} \rightarrow \mathbb{C}^{\mathcal{K}_{N}}, \: \: \mathcal{\hat{G}}_{N}(\hat{v}) = \mathcal{F}^{\alpha}(\mathcal{G}_{N}(\mathcal{F}^{-\alpha}(\hat{v})))
\end{equation}

we can show that,
\begin{equation}
    \mathcal{G}_{N} = \mathcal{F}^{-\alpha} \circ \mathcal{\hat{G}}_{N} \circ (\mathcal{F}^{\alpha} \circ \mathcal{P}_{N})
\end{equation}

Now with above decomposition, we can construct \cono{} by separately approximating each individual terms $\mathcal{F}^{-\alpha}$, $\mathcal{\hat{G}}_{N}$ and $\mathcal{F}^{\alpha} \circ \mathcal{P}_{N}$.

\textbf{Approximating $\mathcal{\hat{G}}_{N}$}: For $\forall \varepsilon \geq 0$ and $\mathcal{\hat{G}}_{N}: \mathbb{C}^{\mathcal{K}_{N}} \rightarrow \mathbb{C}^{\mathcal{K}_{N}}$, we have $l$ \cono{} layers $\mathcal{L}_{l} \circ \mathcal{L}_{l-1} \circ ... \circ \mathcal{L}_{1}$ which satisfy:

\begin{equation}
    ||\mathcal{\hat{G}}_{N}(\hat{v})(x) - (\mathcal{L}_{l} \circ \mathcal{L}_{l-1} \circ ... \circ \mathcal{L}_{1})(\hat{w})(x)||_{L^2} \leq \varepsilon, \:\: \forall v \in K, \forall x \in \mathcal{T}^{d}_{\alpha}
\end{equation}
where, $\hat{w}: \mathcal{T}^{d}_{\alpha}, x \rightarrow \hat{v}$ is constant function defined on $\mathcal{T}^{d}_{\alpha}$.

From \cono{} layer definition, we have, 

\begin{equation}
    \mathcal{L}_{l}(v)(x) = Wv(x) + \mathcal{F}^{-\alpha}(K_{l}(k)\mathcal{F}^{\alpha}(\mathcal{P}_{N}v)(k))
\end{equation}

To approximate $\mathcal{\hat{G}}_{N}$, we can use \textit{Universal Approximation Theorem for CVNNs} \cite{voigtlaender2023universal} by setting $K_{l}(k)$ to be identity.
Then,

\begin{equation}
    \mathcal{F}^{-\alpha}(K_{l}(k)\mathcal{F}^{\alpha}(\mathcal{P}_{N}v)(k)) \approx \mathcal{P}_{N}(v), \:\: \forall l, 
\end{equation}

then we have,
\begin{equation}
    \mathcal{L}_{l}(v)(x) = Wv(x) + \mathcal{P}_{N}(v)(x)
\end{equation}

And then using the Universal Approximation Theorem for CVNNs will guarantee that $\mathcal{L}_{l} \circ \mathcal{L}_{l-1} \circ ... \circ \mathcal{L}_{1}$ can approximate $\mathcal{\hat{G}}_{N}$ to any desired precision.

\textbf{Approximating $\mathcal{F}^{\alpha} \circ \mathcal{P}_{N}$}: For any $\varepsilon \geq 0$ there exists $l \geq 0$ and $\mathcal{L}_{l} \circ \mathcal{L}_{l-1} \circ ... \circ \mathcal{L}_{1} \circ \mathcal{P}$ that satisfy following:
\begin{equation}
    ||\mathcal{F}^{\alpha}(\mathcal{P}_{N}v) - (\mathcal{L}_{l} \circ \mathcal{L}_{l-1} \circ ... \circ \mathcal{L}_{1} \circ \mathcal{P})(v)(x)||_{L^2} \leq \varepsilon, \:\: \forall v \in K, \forall x \in \mathcal{T}^{d}
\end{equation}

Let's define $\mathcal{R}(k, x) = e_{\alpha}(k)e_{\alpha}(k, x)$ which constitutes the orthonormal basis for FrFT, which we call as fractional fourier basis.

Firstly we will show that there exists $\mathcal{N} = \mathcal{L}_{l} \circ \mathcal{L}_{l-1} \circ ... \circ \mathcal{L}_{1} \circ \mathcal{P}$ that satisfies:
\begin{equation}
    \begin{cases}
        ||\mathcal{N}(v)_{1, k} - \mathcal{P}_{N}v(x).Re(\mathcal{R}(k, x)) ||_{L_2} < \varepsilon \\
        ||\mathcal{N}(v)_{1, k} - \mathcal{P}_{N}v(x).Im(\mathcal{R}(k, x)) ||_{L_2} < \varepsilon
    \end{cases}
\end{equation}

To construct $\mathcal{N}$, we define the lifting operator $\mathcal{P}$ and use position embedding as follows:

\begin{equation}
    \mathcal{P}(v) = R(v(x) = \{ v(x), Re(\mathcal{R}(k, x)), v(x), Im(\mathcal{R}(k, x))\}_{k \in \mathcal{K}_{N}}
\end{equation}

where trigonometric polynomial of order $\alpha$ are directly embedded in \cono{} layer. $\mathcal{P}$ lifts the range of function from $\mathbb{R}^{d_{a}}$ to $\mathbb{R}^{d_{u}}$. Then, by leveraging the Universal Approximation Theorem for CVNNs where we concatenate the $\{ v.Re(\mathcal{R}(k, x)), v.Im(\mathcal{R}(k, x))\}_{k \in \mathcal{K}_{N}}$, we can have multiple layers satisfies: 
\begin{equation}
    \scriptstyle (\mathcal{L}_{l} \circ ... \circ \mathcal{L}_{1})(\{ v(x), Re(\mathcal{R}(k, x)), v(x), Im(\mathcal{R}(k, x))\}_{k \in \mathcal{K}_{N}}) \approx \{ v.Re(\mathcal{R}(k, x)), v.Im(\mathcal{R}(k, x))\}_{k \in \mathcal{K}_{N}}, \:\: \forall v \in K
\end{equation}

And can achieve the desired precision by adjusting the width and depth of the layers.
We next note that,
\begin{equation}
    \mathcal{P}_{N}v(x) = \sum_{k \in \mathcal{K}_{N}} \hat{v}_{k} K_{-\alpha}(k, x)  
\end{equation}

where $\hat{v}_{k}$ denotes the $k^{th}$ coefficient of FrFT. Then by definition of FrFT in Eq \ref{eq:frft}:

\begin{equation}
    \mathcal{F}^{\alpha}(v)(0) = \int_{\mathcal{T}^{d}_{\alpha}} v(x) A_{\alpha} e_{\alpha}(x) dx = Re(\hat{v}_{0}) + Im(\hat{v}_{0})
\end{equation}

\begin{equation}
    \mathcal{F}^{\alpha}(v.\mathcal{R}(k, x))(0) = \int_{\mathcal{T}^{d}_{\alpha}} v(x) A_{\alpha} e_{\alpha}(x) e_{\alpha}(k)e_{\alpha}(k, x) dx = Re(\hat{v}_{k}) + Im(\hat{v}_{k})
\end{equation}

Thus in layer $\mathcal{L}_{l + 1}$:
\begin{equation}
    \begin{cases}
        K_{l+1}(k) = 0 & \text{if} \:\: k = 0 \\
        K_{l+1}(k) = -Id & \text{if} \:\: k \neq 0 \\
    \end{cases}
\end{equation}

And then we have:
\begin{equation}
    \scriptstyle \mathcal{F}^{-\alpha}(K_{l+1}(k). \mathcal{F}^{\alpha}(\{ v.Re(\mathcal{R}(k, x)), v. Im(\mathcal{R}(k, x))\}_{k \in \mathcal{K}_{N}})) = \{Re(\hat{v}_{k}) - v.Re(\mathcal{R}(k, x)), Im(\hat{v}_{k}) - v. Im(\mathcal{R}(k, x))\}_{k \in \mathcal{K}_{N}}
\end{equation}

Finally, we can set $W_{l+1}$ in layer $\mathcal{L}_{l+1}$ to be identity we get,

\begin{equation}
    \mathcal{L}_{l+1}(\{ v.Re(\mathcal{R}(k, x)), v. Im(\mathcal{R}(k, x))\}_{k \in \mathcal{K}_{N}})) = \{ Re(\hat{v}_{k}), Im(\hat{v}_k)\}_{k \in \mathcal{K}_{N}} = \mathcal{F}^{\alpha}(\mathcal{P}_{N}v)
\end{equation}

Which is the desired approximation.

\textbf{Approximating $\mathcal{F}^{-\alpha}$}: For any $\varepsilon$, we have $l \geq 0$ and $\mathcal{Q} \circ \mathcal{L}_{l} \circ \mathcal{L}_{l-1} \circ ... \circ \mathcal{L}_{1}$ that satisfies:

\begin{equation}
    ||\mathcal{F}^{-\alpha}(\hat{v}) - (\mathcal{Q} \circ \mathcal{L}_{l} \circ \mathcal{L}_{l-1} \circ ... \circ \mathcal{L}_{1})(v)||_{L^{2}} \leq \varepsilon, \:\: \forall v \in K
\end{equation}

where $\hat{v} \in \mathbb{C}^{\mathcal{K}_{N}}$ is the truncated fractional fourier coefficient.

We can use the previous steps here as well. We can construct $Re(\hat{v}_{k}).\mathcal{R}(k, x)$ and $Im(\hat{v}_{k}).\mathcal{R}(k, x)$ using $\mathcal{L}_{l} \circ \mathcal{L}_{l-1} \circ ... \circ \mathcal{L}_{1}$. We can construct projection operator $\mathcal{Q}$ to recover the original function.

\begin{equation}
    \mathcal{Q}(\{ \hat{v}_{k}.\mathcal{R}(k, x)\}_{k \in \mathcal{K}_{N}}) = \sum_{k \in \mathcal{K}_{N}} Re(\hat{v}_{k}).\mathcal{R}(k, x) - Im(\hat{v}_{k}).\mathcal{R}(k, x)= v(x) 
\end{equation}

Therefore, it completes our proof.

Using the above approximation for $\mathcal{G}_{N}$ we can approximate the operator to desired precision. Thus establishing the Universal Approximation Theorem for \cono{}.

\section{Details for the Benchmark} \label{Appendix C}
\subsection{Details for benchmarks tasks}
The following sections comprehensively elucidate details for benchmark tasks.

\begin{figure*}[!htb]
    \centering
    \includegraphics[width=1.0\linewidth]{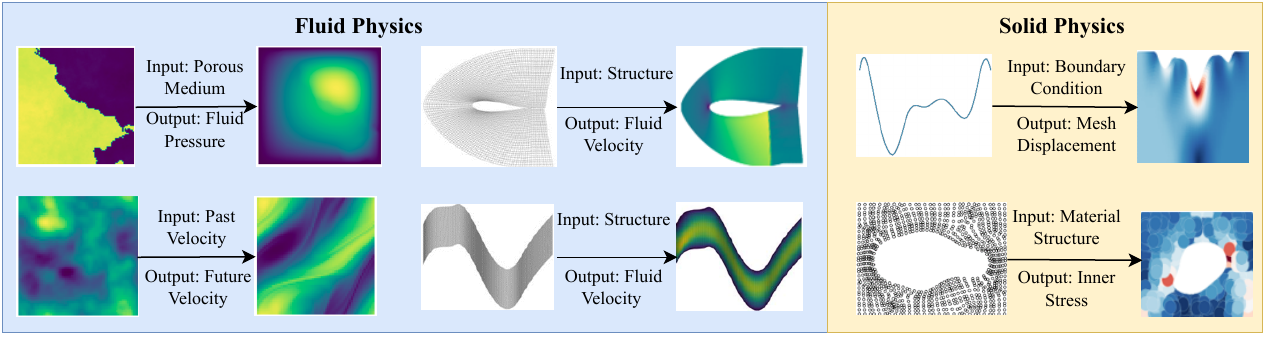}
    \caption{\label{fig:figure8} \small The diagram presents an overview of the Operator learning task applied to distinct Partial Differential Equations (PDEs) classified as Fluid Physics and Solid Physics. \textbf{(Top)} showcases three specific PDEs: Darcy Flow, Airfoil, and Plasticity. \textbf{(Bottom)}, an additional set of three PDEs is featured: Navier Stokes, Pipe, and Elasticity.}
\end{figure*}

\begin{table}[hbt!]
    \centering
    \caption{\label{tab:dataset_details} \small Detailed benchmark tasks provide a thorough and systematic exploration of each task's specific characteristics used in experimentation.}
    \resizebox{0.8\textwidth}{!}{
    \begin{tabular}{l|c|c|c|c}
        \toprule
        \textbf{Dataset} & \textbf{Geometry} &\textbf{Task}            & \textbf{Input} & \textbf{Output} \\
        \midrule
        \midrule
        ELASTICITY-P & Point Cloud & Estimate Stress & Material Structure & Inner Stress \\
        ELASTICITY-G & Regular Grid &Estimate Stress & Material Structure & Inner Stress              \\
        PLASTICITY         & Structured Mesh & Model Deformation        & Boundary Condition       & Mesh Displacement         \\
        NAVIER-STOKES      & Regular Grid & Predict Future           & Past Velocity            & Future Velocity           \\
        AIRFOIL            & Structured Mesh & Estimate Velocity        & Structure                & Fluid Velocity            \\
        PIPE               & Structured Mesh & Estimate Velocity        & Structure                & Fluid Velocity            \\
        DARCY              & Regular Grid & Estimate Pressure        & Porous Medium            & Fluid Pressure            \\
        \bottomrule
        \end{tabular}}
\end{table}

\subsection{Description of Datasets}
Table \ref{tab:dataset_details} and Fig. \ref{fig:figure8} comprehensively present the benchmark details. The categorization of generation details is presented according to the governing partial differential equations (PDEs) as follows:

\textbf{Elasticity-P and Elasticity-G Dataset \cite{li2022fourier}:} 
The dataset is designed to evaluate internal stress within an incompressible material characterized by an arbitrary void at its center, subject to external tension. The material's structural configuration constitutes the input, and the resulting internal stress is the output. Notably, two distinct approaches are employed in modeling the material's geometry: Elasticity-P utilizes a point cloud comprising 972 points. Elasticity-G represents the data on a structured grid with dimensions $41 \times 41$, obtained through interpolation from the Elasticity-P dataset.

\textbf{Plasticity Dataset \cite{li2022fourier}:} This dataset addresses the plastic forging scenario, wherein a die with an arbitrary shape impacts a plastic material from above. The input to the benchmark is characterized by the die's shape, encoded in a structured mesh. The benchmark aims to predict the deformation of each mesh point over the subsequent 20 timesteps. The structured mesh employed has a resolution of $101 \times 31$.

\textbf{Navier Stokes Dataset \cite{li2020fourier}:} 2D Navier-Stokes equation mathematically describes the flow of a viscous, incompressible fluid in vorticity form on the unit torus as follows:

\begin{equation}
    \begin{split}
        \partial_t w(x,t) + u(x,t) \cdot \nabla w(x,t) &= \nu \Delta w(x,t) + f(x), \quad x \in (0,1)^2, \, t \in (0,T] \\
        \nabla \cdot u(x,t) &= 0, \quad x \in (0,1)^2, \, t \in [0,T] \\
        w(x,0) &= w_0(x), \quad x \in (0,1)^2
    \end{split}
\end{equation}

where, $u$ represents the velocity field, $w = \nabla \times u$ is the vorticity, $w_0$ is the initial vorticity, $\nu$ is the viscosity coefficient, and $f$ is the forcing function. 
In this dataset, the viscosity ($\nu$) is fixed at $10^{-5}$, and the 2D field has a resolution of $64 \times 64$. Each sample within the dataset comprises 20 consecutive frames. The objective is to predict the subsequent ten frames based on the preceding ten.

\textbf{Pipe Dataset \cite{li2022fourier}:} This dataset focuses on the incompressible flow through a pipe. The governing equations are Equation \ref{eq:governing}:

\begin{equation}
    \nabla \cdot \mathbf{U} = 0,
\end{equation}

\begin{equation}
    \frac{\partial \mathbf{U}}{\partial t} + \mathbf{U} \cdot \nabla \mathbf{U} = \mathbf{f}^{-1} \frac{1}{\rho} \nabla p + \nu \nabla^2 \mathbf{U}. 
    \label{eq:governing}
\end{equation}

The dataset is constructed on a geometrically structured mesh with a $129 \times 129$ resolution. We employ the mesh structure as input data for experimental purposes, with the output being the horizontal fluid velocity within the pipe.

\textbf{Airfoil Dataset \cite{li2022fourier}}: The dataset pertains to transonic flow over an airfoil. Due to the negligible viscosity of air, the viscous term $\nu \nabla^2U$ is omitted from the Navier-Stokes equation. Consequently, the governing equations for this scenario are expressed as follows:

\begin{equation}
    \frac{\partial \rho f}{\partial t} + \nabla \cdot (\rho f U) = 0
\end{equation}

\begin{equation}
    \frac{\partial (\rho f U)}{\partial t} + \nabla \cdot (\rho f UU + pI) = 0
\end{equation}

\begin{equation}
    \frac{\partial E}{\partial t} + \nabla \cdot ((E + p)U) = 0, \label{eq:navier-stokes}
\end{equation}

where $\rho f$ represents fluid density, and $E$ denotes total energy. The data is on a structured mesh with dimensions $200 \times 50$, and the mesh point coordinates are utilized as inputs. The corresponding output is the Mach number at each mesh point.

\textbf{Darcy Flow Dataset \cite{li2020fourier}}: It represents the flow through porous media. 2D Darcy flow over a unit square is given by 

\begin{equation}
    \nabla \cdot (a(x) \nabla u(x)) = f(x), \quad x \in (0,1)^2, \label{eq:darcy}
\end{equation}

\begin{equation}
    u(x) = 0, \quad x \in \partial(0,1)^2. \label{eq:bc}
\end{equation}

where $a(x)$ is the viscosity, $f(x)$ is the forcing term, and $u(x)$ is the solution. This dataset employs a constant value of forcing term $F(x)=\beta$. 
Further, Equation ~\ref{eq:darcy} is modified in the form of a temporal evolution as

\begin{equation}
    \partial_t u(x,t) - \nabla \cdot (a(x) \nabla u(x,t)) = f(x), \quad x \in (0,1)^2, \label{eq:temporal}
\end{equation}

In this dataset, the input is represented by the parameter $a$, and the corresponding output is the solution $u$. The dataset comprises samples organized on a regular grid with a resolution of $85 \times 85$.

\section{Implementation Details} \label{Appendix D}
The following section provides a comprehensive overview of the training and testing samples, including details about the shapes of input and output tensors.

\begin{table}[htb!]
    \centering
    \caption{\label{tab:implementation} \small Training details for benchmark datasets. The input-output resolutions are presented in the shape of (temporal, spatial, variate). The symbol "/" indicates dimensions excluded.}
    \resizebox{0.8\textwidth}{!}{
    \begin{tabular}{l|c|c|c|c}
    \toprule
    \textbf{Dataset} & \textbf{Training Samples} & \textbf{Testing Samples} & \textbf{Input Tensor} & \textbf{Output Tensor} \\
    \midrule
    \midrule
        ELASTICITY-P  & 1000 & 200 & (/, 972, 2) & (/, 972, 1)   \\
        ELASTICITY-G  & 1000 & 200 & (/, 41 $\times$ 41, 1)  & (/, 41 $\times$ 41, 1) \\
        PLASTICITY    & 900  & 80  & (/, 101 $\times$ 31, 2) & (20, 101 $\times$ 31, 4) \\
        NAVIER-STOKES & 1000 & 200 & (10, 64 $\times$ 64, 1)   & (10, 64 $\times$ 64, 1)  \\
        AIRFOIL       & 1000 & 100 & (/, 200 $\times$ 50, 2)   & (/, 200 $\times$ 50, 1) \\
        PIPE          & 1000 & 200 & (/, 129 $\times$ 129, 2)  & (/, 129 $\times$ 129, 1) \\
        DARCY         & 1000 & 200 & (/, 85 $\times$ 85, 1) & (/, 85 $\times$ 85, 1)  \\
    \bottomrule
    \end{tabular}}
\end{table}

\subsection{Training Details}
Table \ref{tab:implementation} presents a comprehensive overview of the experimental setup, including details regarding the training and testing split and the shapes of the input and output tensors. This information is crucial for understanding the specific configurations employed in our experimentation process. In training \cono{}, we need to initialize the fractional orders, which are learned in the same way as other matrices that are part of our network using optimizers such as Adam. Notably, fractional orders can vary across axes, and there is no requirement for uniformity initialization of fractional orders across different axes. Furthermore, we conducted each experiment five times and observed that the standard deviation falls within the ranges of 0.0003 for Darcy, Airfoil, and Pipe datasets, 0.01 for Navier Stokes, 0.002 for Elasticity-P and Elasticity-G, and 0.0002 for Plasticity.

\subsection{Discrete Implementation of Fractional Fourier Transform}
The discrete implementation of the Fractional Fourier Transform (FrFT) is essential for \cono{}. In \cono{}, we utilize a matrix multiplication-based discrete FrFT. This approach leverages the spectral expansion of the fractional integral kernel using a complete set of eigenfunctions of the FrFT, which are Hermite-Gaussian functions \cite{839980}. We have used the following PyTorch based discrete implementation of FrFT: \href{https://github.com/tunakasif/torch-frft?tab=readme-ov-file#trainable-fractional-fourier-transform}{Torch FrFT}.

\subsection{Complex Neural Networks (CVNNs)}
\begin{wrapfigure}{r}{0.5\textwidth}
    \centering
    \includegraphics[width=0.3\linewidth]{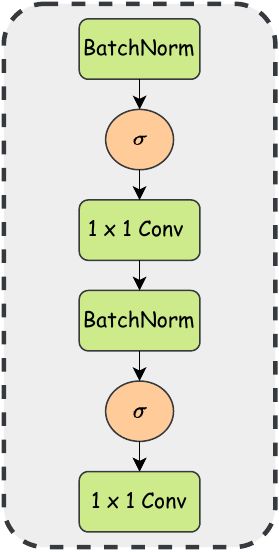}
    \caption{\label{fig:figure9} \small Complex Residual Network used in \cono{} for learning the complex part for CVNNs.}
\end{wrapfigure}

Given that all datasets are inherently real-valued, we designed a neural network optimized for real-valued operations within the complex domain. Using Complex-Valued Neural Networks (CVNNs), the network effectively handles complex data components. It converts real-valued data to complex representations through a residual block, as shown in Fig. \ref{fig:figure9}. This architecture allows the neural network to process the real-valued datasets in the complex domain efficiently. Complex-valued backpropagation was implemented using the Wirtinger calculus \cite{amin2011wirtinger}, which generalizes the notion of complex derivatives and trains complex-valued neural networks (generalized complex chain rule for real-valued loss function) ~\cite{barrachina2023theory, chiheb2017deep}. 
If $L$ is a real-valued loss function and $z$ is a complex variable such that $z = x + iy$ where $x, y \in \mathbb{R}$: 

\begin{equation}
    \nabla_{z} L = \frac{\partial L}{\partial z} = \frac{\partial L}{\partial x} + i \frac{\partial L}{\partial y} = \frac{\partial L}{\partial (\text{Re}(z))} + i \frac{\partial L}{\partial (\text{Im}(z))} = (\nabla_{Re(z)} L + i (\nabla_{Im(z)} L)
\end{equation}

We have used Pytorch to build blocks for \cono{} based on the following GitHub repositories: 
\begin{enumerate}
    \item \href{https://github.com/ChihebTrabelsi/deep\_complex\_networks}{Deep Complex Network}
    \item \href{https://github.com/soumickmj/pytorch-complex}{Pytorch Complex}
\end{enumerate}

\subsection{Hyperparameters}
\begin{wraptable}{r}{3cm}
    \centering
    \caption{\small Table detailing the hyperparameters used in \cono{} in Darcy Flow Benchmark.}
    \label{tab:hyperparameters}
    \resizebox{\linewidth}{!}{ 
    \begin{tabular}{l|c}
        \toprule
        \textbf{Parameters} & \textbf{Values} \\
        \midrule
        \midrule
        Learning Rate & 0.001 \\
        Batch Size & 20 \\
        Latent Dim  & 64 \\
        Padding & 11 \\
        Gamma & 0.5 \\
        Step-Size & 100 \\ 
        \bottomrule
    \end{tabular}}
\end{wraptable} 

This section details the hyperparameter values used in \cono{} for the Darcy Flow dataset. We will specify the settings and configurations, such as learning rates, batch sizes, the number of layers, activation functions, and any other relevant parameters that were employed to optimize the performance of \cono{} on this dataset as shown in Table \ref{tab:hyperparameters}. Further, as shown in Fig. \ref{fig:cono} we initialize $\alpha = 1$ and $\alpha' = 0.5$. We further used pointwise convolution in $R^{\alpha'}$ and linear transformation in $R^{\alpha}$ by truncating the higher frequency similar as FNO.

\subsection{Mitigation of Aliasing}
In Neural Operator learning, the utilization of non-linear operations, such as non-linear pointwise activations, can introduce high-frequency components into the output signal. The manifestation of aliasing induced by nonlinearity can result in distortion of symmetry inherent in the physical signal, leading to undesirable effects. Additionally, the pursuit of translational invariance, a key characteristic in neural operators, becomes susceptible to degradation due to aliasing.
We propose a two-step process to address the challenges of aliasing errors within the continuous equivariance paradigm. Firstly, before applying any activation function, we employ an upsampling operation on the input function, exceeding its frequency bandwidth. Subsequently, a non-linear operation is applied to the upsampled signal, followed by a sinc-based filter and downsampling. Incorporating the sinc-based low-pass filter effectively attenuates higher frequency components in the output signal. This method enhances the accuracy of neural network predictions by minimizing distortions caused by aliasing, which is especially critical in applications dealing with complex signal-processing tasks. Although we experimentally haven't found much improvement in performance using alias-free activation.


\section{Prediction Visualization} \label{Appendix E}
\begin{figure*}[htb!]
    \centering
    \includegraphics[width=1.0\linewidth]{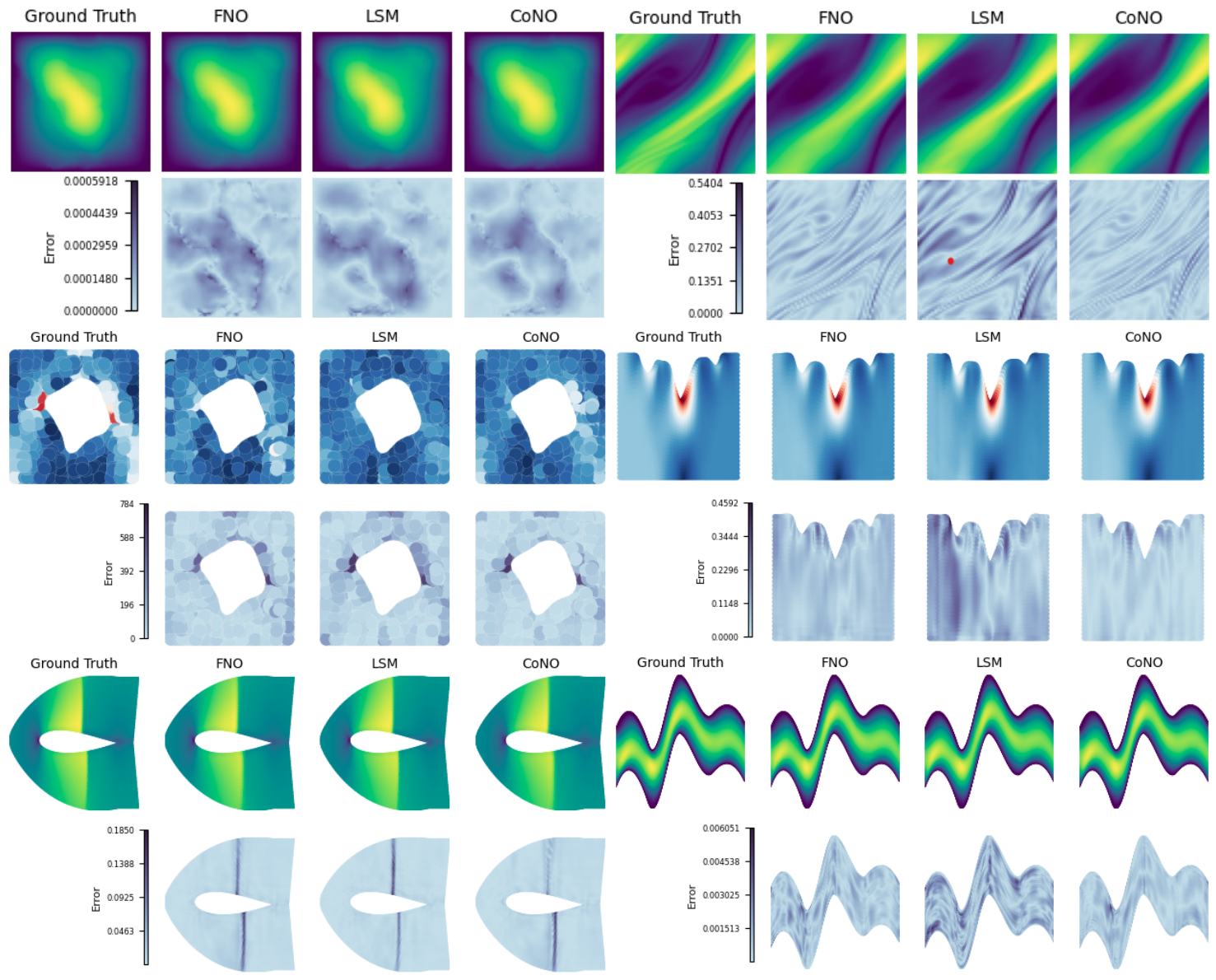}
    \caption{\label{fig:figure10} \small \textbf{(Top)} Showcase of Darcy Flow \textbf{(Left)} and Navier Stokes \textbf{(Right)}. \textbf{(Middle)} Showcase of Airfoil \textbf{(Right)} and Pipe \textbf{(Left)}. \textbf{(Bottom)} Showcase of Plasticity \textbf{(Left)} and Elasticity \textbf{(Right)}. For Comparison of predicted output, we have plotted the heatmap of the absolute value of the difference between Ground Truth and Prediction.}
\end{figure*}

As illustrated in Fig. \ref{fig:figure10}, the performance exhibited by \cono{} in predictive tasks surpasses that of other benchmark datasets, notably outperforming the state-of-the-art operator LSM. This superiority is particularly evident in time-dependent and independent partial differential equations (PDEs) scenarios. \cono{} showcases enhanced predictive accuracy and significantly reduces artifacts. These compelling findings underscore the efficacy of \cono{} as a robust solution for a diverse range of PDE applications, marking a significant advancement in scientific machine learning.

\section{Experimental Results} \label{Appendix F}

The following section provides an exhaustive examination of the results obtained from multiple tasks associated with the operator. These tasks encompass a broad spectrum, including different resolutions, resilience to noise, performance in out-of-generalization tasks, data efficiency, and training stability. The comprehensive analysis presented in this section aims to offer a detailed insight into the performance and capabilities of the operator across a range of critical aspects, contributing to a nuanced understanding of its practical utility and effectiveness in diverse scenarios. 
All baseline models are instantiated and implemented utilizing the official source code.

\subsection{Performance Under Different Resolution} \label{sec:resolution}
To assess the efficacy of our operator at diverse resolutions, we conducted experiments on Darcy and Navier-Stokes problems across varying spatial resolutions. Notably, \cono{} consistently outperformed other operators across these resolutions, demonstrating its superior effectiveness. The comparative results, as presented in Table \ref{tab:resolutiondarcy} and Table \ref{tab:resolutionns}, unequivocally highlight the robust performance and versatility of \cono{} in addressing challenges associated with different spatial resolutions in the context of both Darcy and Navier-Stokes scenarios.

\begin{table}[htb!]
    \centering
    \caption{\label{tab:resolutiondarcy} \small Neural Operators performance comparison on Darcy Flow under Different Resolutions.}
    \resizebox{0.8\textwidth}{!}{
    \begin{tabular}{l|cccccccc}
        \toprule
        \textbf{Resolution} & \textbf{UNET} & \textbf{FNO} & \textbf{MWT} & \textbf{UNO} & \textbf{FFNO} & \textbf{HTNET} & \textbf{LSM} & \textbf{\cono{}}\\
        \midrule
        \midrule
        32x32    & 0.0059 & 0.0128 & 0.0083 & 0.0148 & 0.0103 & 0.0058 & 0.0049 & \textbf{0.0048}\\
        64x64    & 0.0052 & 0.0067 & 0.0078 & 0.0079 & 0.0064 & 0.0046 & 0.0042 & \textbf{0.0036}\\
        128x128  & 0.0054 & 0.0057 & 0.0064 & 0.0064 & 0.0050 & 0.0040 & 0.0038  & \textbf{0.0034}\\
        256x256  & 0.0251 & 0.0058 & 0.0057 & 0.0064 & 0.0051 & 0.0044 & 0.0043 & \textbf{0.0039}\\
        512x512  & 0.0496 & 0.0057 & 0.0066 & 0.0057 & 0.0042 & 0.0063 & 0.0039 & \textbf{0.0035}\\
        1024x1024 & 0.0754 & 0.0062 & 0.0077 & 0.0058 & 0.0069 & 0.0163 & 0.0050 & \textbf{0.0044}\\
        \bottomrule
    \end{tabular}}
\end{table}

\begin{table}[htb!]
    \centering
    \caption{\label{tab:resolutionns} \small Neural Operators performance comparison on the Navier-Stokes Benchmark Under Different Resolutions. "/" indicates poor $l2$ error performance.}
    \resizebox{0.8\textwidth}{!}{
    \begin{tabular}{l|cccccccc}
        \toprule
        \textbf{Resolution} & \textbf{UNET} & \textbf{FNO} & \textbf{MWT} & \textbf{UNO} & \textbf{FFNO} & \textbf{HTNET} & \textbf{LSM} & \textbf{\cono{}}\\
        \midrule
        \midrule
        64×64   & 0.1982 & 0.1556 & 0.1541 & 0.1713 & 0.2322 & 0.1847 & 0.1535 & \textbf{0.1287}\\
        128×128 & /     & 0.1028 & 0.1099 & 0.1068 & 0.1506 & 0.1088 & 0.0961 & \textbf{0.0817}\\
        \bottomrule
    \end{tabular}}
\end{table}

\subsection{Robustness to Noise}\label{sec:noise}
To evaluate the robustness of our operator in the presence of noisy training input, we systematically conducted experiments across varying input noise levels. Our objective was to comprehensively understand the impact of noise on the performance of the proposed operator. Remarkably, our findings revealed that \cono{}, even when subjected to 0.1\% data noise, consistently outperformed LSM—the best-performing operator trained without exposure to noisy data. This result is a compelling confirmation of the enhanced robustness exhibited by our proposed operator under challenging conditions involving noisy training inputs, as highlighted in Table \ref{tab:noisytraining}.

We observe relative error increases of 22\%, 47\%, and 28\% for CoNO, LSM, and FNO, respectively. Here, CoNO outperforms all the models. Furthermore, under 0.5\% data noise conditions, our findings indicate relative error increases of 62\%, 186\%, and 45\% for CoNO, LSM, and FNO, respectively. Here, it performs significantly better than LSM but poorer than FNO. However, it should be noted that the absolute performance of CoNO in relation to FNO is considerably better. Thus, it can be argued that CoNO is robust against noise at least as much as FNO, if not better and better LSM, which is the SOTA model.

\begin{table}[htb!]
    \centering
    \caption{\label{tab:noisytraining} \small Neural Operators performance comparison on Darcy Flow under different noise Levels in the training dataset. "/" indicates poor $\ell_2$ error performance.}
    \resizebox{0.8\textwidth}{!}{
    \begin{tabular}{l|cccccccc}
        \toprule
        \textbf{Noise \%} & \textbf{UNET} & \textbf{FNO} & \textbf{MWT} & \textbf{UNO} & \textbf{FFNO} & \textbf{HTNET} & \textbf{LSM} & \textbf{\cono{}}\\
        \midrule
        \midrule
        0 \%    & 0.0080 & 0.0108 & 0.0082 & 0.0113 & 0.0083 & 0.0079 & 0.0065 & \textbf{0.0050}\\
        0.001 \%    & 0.0094 & 0.0114 & 0.0089 & 0.0115 & 0.0089 & 0.0087 & 0.0085 & \textbf{0.0056}\\
        0.01 \%    & 0.0105 & 0.0137 & 0.0097 & 0.0121 & 0.0095 & 0.0097 & 0.0087 & \textbf{0.0058}\\
        0.1 \%  & 0.0113 & 0.0139 & 0.0125 & 0.0126 & 0.0106 & 0.0124 & 0.0096 & \textbf{0.0061}\\
        0.5 \%    & / & 0.0157 & 0.0135 & 0.0138 & 0.0125 & 0.0148 & 0.0186 & \textbf{0.0081}\\
        \bottomrule
    \end{tabular}}
\end{table}

\subsection{Out of Distribution Generalization}\label{sec:out}
In this study, we conducted extensive experiments utilizing the Navier-Stokes dataset, training our model with a viscosity coefficient set to $10^{-5}$. Subsequently, we rigorously assessed the out-of-distribution generalization capabilities by subjecting the trained model to a viscosity coefficient of $10^{-4}$, as depicted in Table \ref{tab:ood generalization}. Our empirical observations consistently demonstrate that the \cono{} model exhibits a notably superior generalization performance, showcasing an impressive increment of $64.3 \%$ compared to the FNO model's performance. Furthermore, our findings underscore the critical importance of capturing latent variable information, a task effectively accomplished by the UNET architecture, particularly exemplified by the Latent Spectral Operator. Significantly, LSM outperforms all other operators, including \cono{}, emphasizing its role in enhancing generalization capabilities in fluid dynamics modeling.
\begin{table}[!htb]
    \centering
    \caption{\label{tab:ood generalization} \small Neural Operators performance comparison on the Navier-Stokes Benchmark Under Out of distribution performance. Model is trained on NS viscosity coefficient $1e^{-5}$ and tested on NS viscosity coefficient $1e^{-4}$. "/" indicates poor $\ell_2$ error performance.}
    \label{tab:out_of_distribution}
    \resizebox{0.8\textwidth}{!}{
    \begin{tabular}{c|cccccccc}
        \toprule
        \textbf{NS Viscosity Coefficient} & \textbf{UNET} & \textbf{FNO} & \textbf{MWT} & \textbf{UNO} & \textbf{FFNO} & \textbf{HTNET} & \textbf{LSM} & \textbf{\cono{}}\\
        \midrule
        \midrule
        $1e^{-5}$  & 0.1982 & 0.1556 & 0.1541 & 0.1713 & 0.2322 & 0.1847 & 0.1535 & \textbf{0.1287}\\
         $1e^{-4}$ & /   & 0.6621 & 0.5864 & 0.5436 & 0.5606 & 0.4888 & \textbf{0.1887} & 0.2321\\
        \bottomrule
    \end{tabular}}
\end{table}

\subsection{Effect of Number of Layers}\label{sec:layers}
We conducted experiments to analyze the relationship between the number of layers and performance, in comparison to FNO, on the Darcy flow dataset. Our findings, presented in the table below, reveal that performance initially improves with increased layers in FNO. However, a significant drop in performance occurs due to the vanishing gradient problem. Importantly, our method consistently avoids the vanishing gradient problem even with an increase in the number of layers as shown in Table \ref{tab:layers}.

\begin{table}[!htb]
    \centering
    \caption{\label{tab:layers} \small Neural Operators performance comparison on Darcy Benchmark under different numbers of layers.}
    \begin{tabular}{c|cccc}
    \toprule
    \textbf{Number of Layers} & 2      & 4      & 6      & 8      \\ \midrule \midrule
    FNO              & 0.0114 & 0.0108 & 0.0087 & 0.0098 \\ 
    CoNO             & \textbf{0.0066} & \textbf{0.0052} & \textbf{0.0053} & \textbf{0.0052} \\ \bottomrule
    \end{tabular}
\end{table}

\subsection{Long Term Prediction} \label{sec:long}
In further substantiating our evidence regarding the enhanced stability of CoNO in long-horizon predictions, we conducted an additional experiment utilizing the Navier Stokes model with a viscosity coefficient of $10^{-4}$. Here, we performed forecasts for the subsequent ten steps based solely on the preceding ten observations and extrapolated the results over ten more steps. The outcomes of this experiment are presented in the tables below. Our findings indicate that although not explicitly trained for predicting the subsequent 20 timestamps, CoNO consistently outperforms LSM and FNO in extrapolating beyond the prediction horizon. This observation underscores the heightened stability and robustness of CoNO in long-horizon prediction tasks as shown in Table \ref{tab:extrapolation}.

\begin{table}[!htb]
    \centering
    \caption{\label{tab:extrapolation} 
    \small A comparative analysis of the performance of Neural Operators on the Navier-Stokes equations, with a viscosity coefficient of $10^{-4}$, involves training models to predict the subsequent 10 timestamps based on the preceding 10 timestamps. Subsequently, the extrapolated results are utilized to forecast the subsequent 10 timestamps.}
    \begin{tabular}{l|cccc}
        \toprule
        \textbf{Neural Operator} & $T = 5$ & $T = 10$ & $T = 15$ & $T = 20$ \\ \midrule \midrule
        FNO             & 0.028   & 0.050    & 0.17     & 0.34     \\
        LSM             & 0.065   & 0.13     & 0.24     & 0.38     \\
        CoNO            & \textbf{0.020}   & \textbf{0.045}    & \textbf{0.14}     & \textbf{0.31}     \\ \bottomrule
    \end{tabular}
\end{table}

\subsection{Data Efficiency} \label{sec:data}
As shown in Table \ref{tab:data_efficiency}, the performance of \cono{} is superior, showcasing its competitive capabilities comparable to the second-best operator LSM when trained on 60\% of the available data. Remarkably, across diverse training dataset ratios, \cono{} consistently surpasses all other operators, emphasizing its remarkable data efficiency compared to state-of-the-art (SOTA) operators. This observation underscores the efficacy and robustness of \cono{} across varying training scenarios, positioning it as a noteworthy solution in the realm of operator-based learning. With a data ratio of 0.6, we observed relative error increases of 34\%, 44\%, and 35\% for CoNO, LSM, and FNO, respectively. Thus, it is evident that our approach consistently outperforms the second-best operator, LSM, across various ratio settings. Further, the absolute performance of CoNO is significantly better than that of FNO.

\begin{table}[!htb]
    \centering
    \caption{\label{tab:data_efficiency} \small Neural Operators performance comparison on Darcy Benchmark under different training dataset ratios. "/" indicates poor $\ell_2$ error performance.}
    \resizebox{0.8\textwidth}{!}{
    \begin{tabular}{l|cccccccc}
        \toprule
        \textbf{Ratio} & \textbf{UNET} & \textbf{FNO} & \textbf{MWT} & \textbf{UNO} & \textbf{FFNO} & \textbf{HTNET} & \textbf{LSM} & \textbf{\cono{}}\\
        \midrule
        \midrule
        0.2    & / & 0.2678 & 0.2854 & 0.2734 & 0.2573 & 0.2564 & 0.2465 & \textbf{0.2234}\\
        0.4    & / & 0.0176 & 0.0165 & 0.0183 & 0.0153 & 0.0145 & 0.0138 & \textbf{0.0105} \\
        0.6  & 0.1234 & 0.0146 & 0.0113 & 0.0153 & 0.0142 & 0.0105 & 0.0094 & \textbf{0.0067}\\
        0.8  & 0.0107 & 0.0122 & 0.0096 & 0.0134 & 0.0095 & 0.0094 & 0.0082 & \textbf{0.0056} \\
        1.0  & 0.0080 & 0.0108 & 0.0082 & 0.0113 & 0.0077 & 0.0079 & 0.0065 & \textbf{0.0050}\\
        \bottomrule
    \end{tabular}}
\end{table}

\subsection{Inference Time Comparison}\label{sec:inference}
The section examines inference time complexity across diverse neural operators applied to the Darcy Flow benchmark dataset within the Fluid Physics field. It aims to elucidate these operators' computational efficiency and efficacy in modeling fluid flow phenomena, as shown in Table \ref{tab:inference}.

\begin{table}[ht]
    \centering
    \caption{\label{tab:inference}
    \small The evaluation of inference time complexity across various neural operators on the Darcy Flow benchmark dataset in the domain of Fluid Physics.}
    \begin{tabular}{c|cccccc}
    \toprule
    \textbf{Neural Operator} & \textbf{UNET} & \textbf{UFNO} & \textbf{FNO} & \textbf{WMT} & \textbf{LSM} & \textbf{CoNO} \\ \midrule \midrule
    \textbf{Inference (s)} & 0.035 & 0.042 & 0.135 & 0.045 & 0.020 & 0.055 \\ 
    \bottomrule
    \end{tabular}
\end{table}

\section{Limitation and Future Work} \label{Appendix G}
To enhance our understanding of \cono{}, it is crucial to explore its mathematical and algorithmic principles thoroughly. We aim to uncover the latent space's learning mechanisms and build a solid theoretical foundation for complex operators. Our research highlights several key challenges that need careful investigation, including refining initialization procedures for fractional orders, designing streamlined architectures for complex neural operators, developing equivariant complex operators, and understanding the role of the Fractional Fourier Transform (FrFT) in the continuous dynamics of complex systems. Additionally, our work raises questions about creating foundational models for Partial Differential Equations (PDEs). These research directions offer opportunities to further understand \cono{} and contribute to the broader field of Scientific Machine Learning (SciML).

\section{Broader Impact} \label{Appendix H}
This paper introduces innovative tools to revolutionize scientific machine learning (SciML). By integrating Complex Neural Networks (CVNNs) and the Fractional Fourier Transform (FrFT) into the Neural Operator framework, we offer a novel approach to addressing the complex challenges of partial differential equations (PDEs), particularly Fractional PDEs, which lack explicit differential forms and are common in natural phenomena \cite{ahmad2020solution, li2018numerical}. Our research has significant implications for diverse fields such as biology, physics, and civil engineering, addressing a crucial scientific problem and paving the way for transformative advancements in interdisciplinary problem-solving. There is no serious ethical issue of the proposed methodology.

\end{document}